# Domain Shift in Echocardiography: Interpretable Quantification and Prediction of Cross-Dataset Left Ventricular Segmentation

Soroush Elyasi[1], Nasim Dadashi Serej[1]*, Julie Wall[2], and Massoud Zolgharni[1]
[1]THRIVE Centre, School of Computing and Engineering, University of West London, London, UK
[2]School of Computing and Engineering, University of West London, London, UK

**Abstract.** Cross-dataset generalisation remains a major barrier to clinical deployment of echocardiographic left ventricular segmentation, yet the sources of this domain shift are rarely disentangled. Using six echocardiographic datasets, we examined whether transfer degradation could be estimated before deployment from handcrafted ultrasound descriptors, VAE latent features, and segmentation-derived latent features. Geometry-aware preprocessing substantially improved several poor cross-dataset transfer cases, suggesting that much of the apparent domain shift reflects field-of-view and framing inconsistencies rather than intrinsic acoustic differences alone. Intensity z-normalisation had little effect on dataset separability, indicating that brightness and contrast were not the dominant shift axis. Transfer degradation was moderately predictable on held-out source-target pairs, and meaningful predictive performance was retained after excluding LV- and fan-specific features, supporting the potential of whole-image descriptors for transfer-risk estimation. The most informative discrepancy measure varied across handcrafted, VAE, and segmentation-derived representations. Manufacturer-related differences could not be reliably disentangled from dataset identity given the limited availability of vendor metadata. Echocardiographic domain shift is therefore structured and measurable, and its impact on segmentation can be partly reduced through geometry-aware preprocessing and anticipated using representation-specific transfer-risk estimation.

**Keywords:** Domain Shift, Medical Image Analysis, Ultrasound Imaging, Echocardiography, Computer Vision, Image Processing, Interpretable Machine Learning

## 1 Introduction

Echocardiography is a non-invasive ultrasound imaging technique used for cardiac assessment, offering a safe and cost-effective means of evaluating heart structure and function [1], [2], [3]. The datasets analysed in this study include images acquired using systems from three major ultrasound manufacturers, GE HealthCare (GE), Philips Healthcare (PH), and Siemens Healthineers (SI). These vendors rely on proprietary imaging pipelines incorporating differences in beamforming, dynamic range compression, speckle reduction, and post-processing algorithms [4], [5], [6], leading to differences in image appearance and statistical properties [3], [7].

Deep learning (DL)-based methods have substantially advanced automated left ventricular (LV) segmentation, enabling scalable segmentation with strong performance [8], [9], [10], [11]. However, despite achieving high accuracy within individual datasets, these models often suffer substantial performance degradation when applied to data from different sources. This decline is driven by domain discrepancies arising from variations in data distribution, noise characteristics, anatomical variability, imaging conditions, scanner types, acquisition protocols, and dataset-specific preprocessing pipelines such as cropping, masking, and image standardisation [3], [12], [13], [14].

Although domain shift is widely recognised, it remains insufficiently understood in echocardiographic LV segmentation. Most existing studies address the problem using model-centric solutions such as domain adaptation or data augmentation, often treating domain shift as a black-box phenomenon [3], [9], [11]. As a result, the underlying sources of domain shift are rarely analysed explicitly, and it remains unclear which factors genuinely affect cross-dataset generalisation, and which are artefacts of preprocessing or dataset construction. A clearer understanding of these factors is essential for developing more robust and efficient models and for distinguishing intrinsic domain differences from preprocessing effects [2], [3].

Another limitation of existing work is the limited use of handcrafted image features, which provide greater interpretability than DL outputs and latent representations. Although less expressive, these features offer valuable insight into underlying image properties and can complement data-driven approaches. Combining handcrafted and learned representations provides a more comprehensive analysis of domain shift. It remains unclear how these different representations capture domain discrepancies and relate to downstream segmentation performance.

To address these limitations, this study develops an interpretable framework for analysing and predicting domain shift in echocardiographic LV segmentation across multiple datasets. Unlike approaches that treat domain shift mainly as a model failure to be corrected after deployment, the proposed framework aims to identify which aspects of the imaging domain explain cross-dataset segmentation degradation. It combines interpretable handcrafted descriptors, including intensity, texture, speckle, image-quality, and geometric features, with unsupervised VAE latent features and segmentation-derived task-specific latent representations. Distribution-based discrepancy measures are then used to quantify source-target differences and determine which representations and distance metrics are most informative for estimating transfer risk.

A central contribution of this work is its controlled multi-setting analysis, which examines the contribution of preprocessing-related variation to observed cross-dataset differences. This is important because apparent domain shift may arise not only from scanner or patient-related variation, but also from preprocessing choices, such as field-of-view differences, anatomical framing, cropping, masking, resizing, image standardisation, and dataset construction. The study also incorporates scanner-vendor information wherever it can be reliably identified from the available data. In particular, samples associated with GE and Philips systems are analysed across datasets to examine whether vendor-related effects can be distinguished from broader dataset-level differences.

Overall, this work provides a unified comparison of image-level handcrafted features, unsupervised latent features, and task-specific segmentation features for domain-discrepancy analysis. It evaluates reliable measures of domain shift, tests whether source-target feature distances can predict cross-dataset transfer performance under held-out validation, and examines whether suitable source datasets can be identified for a given target before deployment. In doing so, the study demonstrates the practical value of interpretable feature-distance analysis for transfer-risk estimation and shows that dataset separability and transfer performance are related but not equivalent. This distinction clarifies why a dataset can appear visually or statistically distinct without necessarily being the most harmful target for segmentation transfer, and why domain separation should not be interpreted as transfer relevance without task-level validation. Fig. 1 presents an overview of the proposed domain-shift analysis workflow.

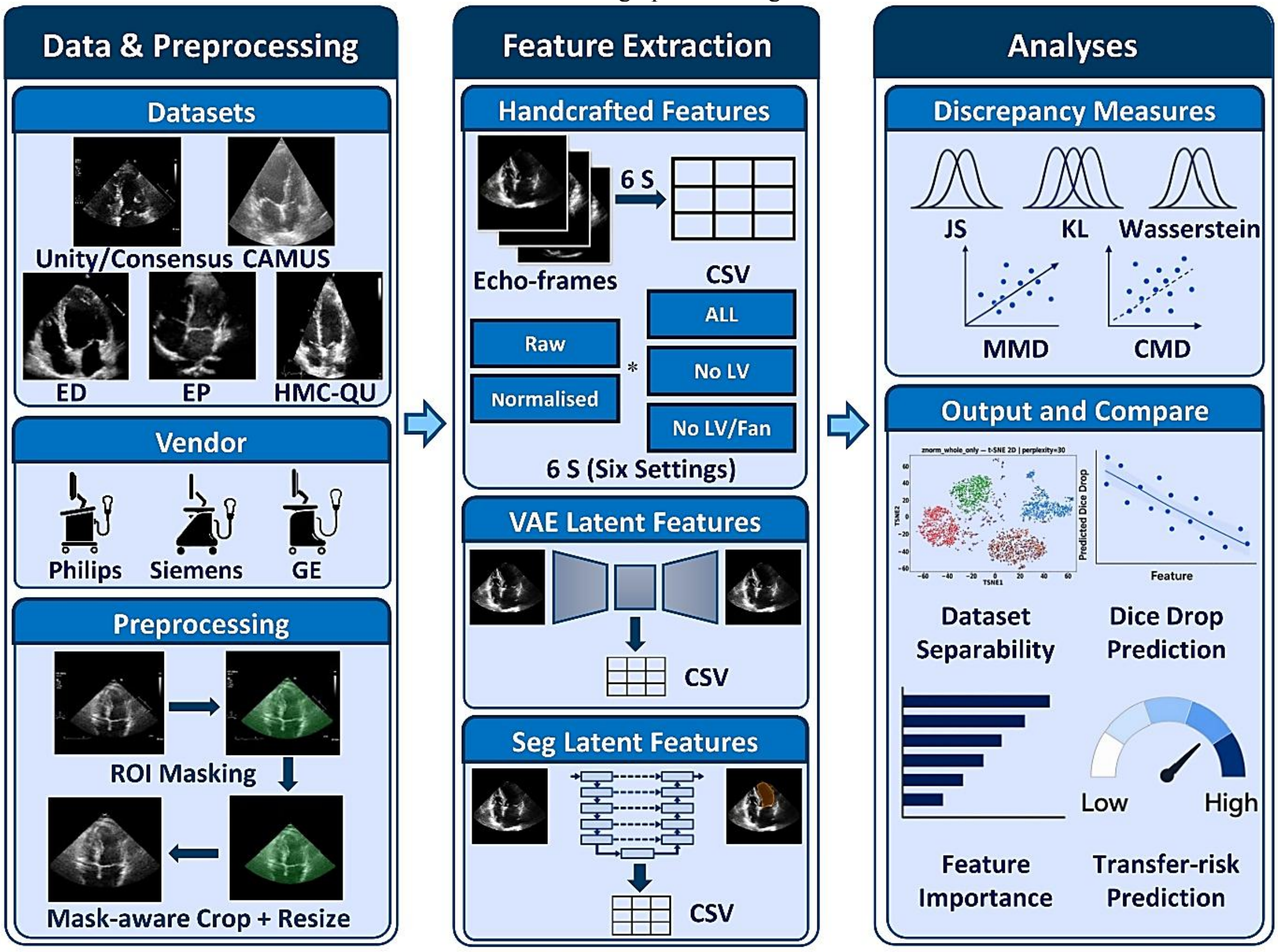

***Fig. 1. Workflow overview of the proposed domain-shift analysis***

## 2 Related Work

Existing research on domain shift in medical imaging can be broadly categorised into several methodological directions, including feature-based analysis, statistical alignment, shift detection, and model-based adaptation.

Feature-based approaches characterise domain shift using interpretable image descriptors. Kushol et al. analysed multi-centre MRI datasets using intensity, noise-related, frequency-domain, wavelet, and texture features, alongside t-SNE and UMAP visualisation, Maximum Mean Discrepancy (MMD), and domain classifiers including Support Vector Machines and Random Forests. Intensity variance, intensity range, and frequency-domain measures were among the most informative features for distinguishing scanner-related domains [15].

Classical texture analysis has also been used to characterise myocardial tissue in echocardiography. Punys et al. identified local mean intensity, histogram shape, local variance, and run-length features as informative for distinguishing healthy from hypertrophic myocardium [16]. Another study found that co-occurrence and run-length measures distinguished normal from contused myocardium, whereas first-order histogram features were not discriminative [17]. They also highlighted the influence of transducer frequency and image-processing settings on echocardiographic texture. Radiomics-based approaches extend this framework through large-scale quantitative feature extraction for echocardiographic analysis. These methods incorporate first-order intensity statistics, such as mean, standard deviation, minimum, and maximum, histogram-based descriptors, such as skewness, kurtosis, entropy, and energy, and higher-order texture features derived from GLCM, GLRLM, NGLDM, and GLZLM [18], [19]. Their experiments show that the high volume of handcrafted and radiomic features is an important consideration in echocardiographic analysis, as broad feature sets may be needed to capture diverse image characteristics. Despite their effectiveness, these methods rely on predefined feature representations and do not explicitly quantify domain discrepancy.

Statistical distribution alignment methods address domain shift by enforcing similarity between feature distributions across domains. Approaches based on MMD, Kullback-Leibler divergence, and Central Moment Discrepancy (CMD) compare statistical properties of feature representations [20], [21]. In particular, CMD aligns higher-order moments, including mean, variance, skewness, and kurtosis, providing a richer description of distributional differences [20]. Classical domain adaptation techniques such as CORAL, Subspace Alignment, Transfer Component Analysis, and Optimal Transport similarly operate by matching statistical properties or transforming feature spaces [21], particularly in unsupervised settings.

A complementary direction focuses on detecting domain shift rather than mitigating it. Methods combining dimensionality reduction, statistical hypothesis testing, and classifier-based approaches have been proposed to identify distribution changes [22]. The Failing Loudly framework demonstrates that Black-Box Shift Detection (BBSD), particularly with soft classifier outputs, effectively captures semantic-level differences and that detectability increases with shift magnitude [22]. Another study shows that low-dimensional projections of bottleneck features in medical segmentation networks preserve domain-relevant information, enabling reliable detection of out-of-distribution data [23].

Empirical studies have demonstrated the impact of scanner variability on deep learning performance. Guo et al. reported cross-scanner performance degradation across MRI, X-ray, and CT, with larger effects observed in MRI and X-ray. Their findings also indicated that noise-based augmentation and the inclusion of target-domain data did not sufficiently mitigate domain shift [24]. Similarly, Yan et al. observed substantial cross-manufacturer degradation in cardiac MRI segmentation, although performance improved following manufacturer adaptation [25].

DL-based domain adaptation methods attempt to address these challenges through representation learning and image transformation. Approaches based on latent-space modelling and adversarial learning aim to learn domain-invariant representations [3], [25]. For instance, Nazari et al. combine a variational autoencoder (VAE) with a Wasserstein generative adversarial network (WGAN) and a dual-attention mechanism to improve robustness to domain variation [3]. Similarly, CycleGAN-based methods perform unpaired image-to-image translation whilst preserving anatomical structure, leading to reported improvements in cross-domain segmentation performance [25].

Perceptual similarity has also been explored for quantifying domain differences. Traditional metrics such as L2 distance, PSNR, and SSIM are often insufficient, whereas deep feature-based similarity measures capture both low-level and high-level structural information by comparing representations across neural network layers [26].

Overall, existing studies address domain shift through feature-based analysis, statistical alignment, and model-based adaptation. However, these approaches remain largely disconnected and are often explored outside the context of echocardiographic ultrasound. Feature-based methods provide interpretability but are rarely linked to downstream segmentation performance, whilst adaptation methods improve generalisation without explicitly explaining domain discrepancies. Detection-based approaches identify the presence of domain shift but do not characterise its underlying causes.

Consequently, there remains a lack of a unified and interpretable framework that systematically compares handcrafted features, learned latent representations, and task-specific features, and explicitly relates domain discrepancies to segmentation performance. Addressing this gap is important for identifying domain characteristics associated with segmentation degradation and informing the development of more reliable models.

# 3 Methods

## 3.1 Datasets

This study used multiple echocardiography datasets acquired from different institutions, scanner vendors, and acquisition protocols to capture a broad range of domain characteristics, as summarised in Table 1 [1], [12], [13], [27], [28], [29], [30]. The selected datasets include images acquired using systems from General Electric (GE), Philips, and Siemens. For each dataset, end-diastolic and end-systolic frames were used in accordance with the provided annotations, whilst preserving the original train/validation/test splits defined by each dataset.

***Table 1. Main dataset summary, splits, vendors, views, and LV-mask availability***

| Characteristics | UN | CS | CA | ED | EP | HMC |
|---|---|---|---|---|---|---|
| Access | Private | Private | Public | Public | Public | Public |
| Train | 2057 | 0 | 1600 | 14930 | 2580 | 288 |
| Validation | 369 | 0 | 200 | 2554 | 336 | 89 |
| Test | 375 | 200 | 200 | 2576 | 368 | 111 |
| Total size | 2801 | 200 | 2000 | 20060 | 3284 | 488 |
| Vendor | PH - GE | PH - GE | GE | PH - SI | PH - SI | PH - GE |
| Views | A4C | A4C | A2C/A4C | A4C | A4C | A2C/A4C |
| ROI Masks | 60 | 40 | 46 | 60 | 60 | 42 |

*Note. A4C, apical four-chamber; A2C, apical two-chamber. ED, EchoNet Dynamic; EP, EchoNet Pediatric; HMC, HMC-QU; CA, CAMUS; CS, Consensus; UN, Unity.*

To provide qualitative insight into inter-dataset variability, representative samples from each dataset are shown in Fig. 2, illustrating the visual variability in image appearance and acquisition characteristics between datasets.

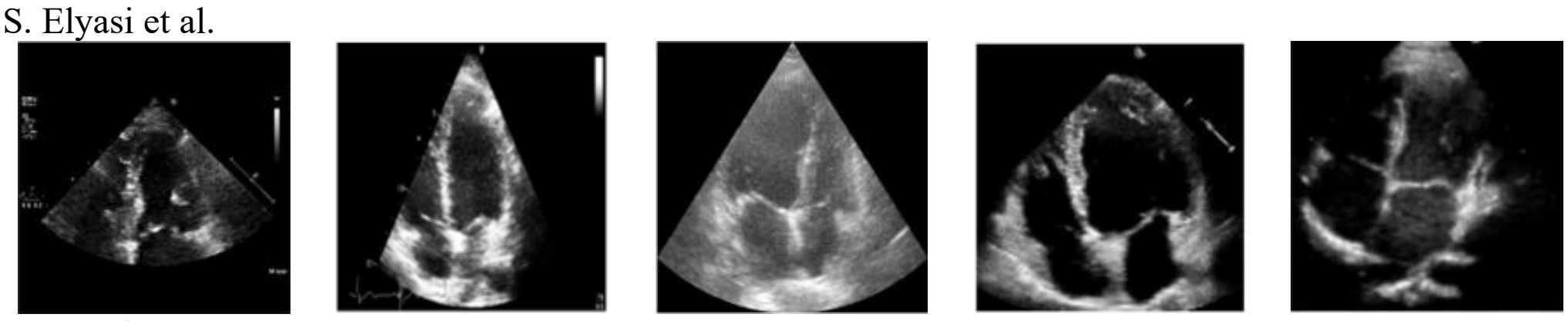

*Unity/Consensus HMC-QU CAMUS ED EP*

***Fig. 2. Representative images from each echocardiography dataset***

In addition, Table 2 summarises the distribution of samples from GE and PH across Unity, Consensus, and CAMUS. Only samples with reliable image-level vendor metadata were included. Consequently, Siemens and datasets without sufficient vendor metadata were excluded from this analysis, resulting in different sample totals from those reported in Table 1.

***Table 2. Vendor distribution across datasets and splits***

| | UN | CS | CA | Total |
|---|---|---|---|---|
| GE - Train | 140 | 0 | 1600 | 1740 |
| GE - Validation | 25 | 0 | 200 | 225 |
| GE - Test | 16 | 2 | 200 | 218 |
| PH - Train | 1701 | 0 | 0 | 1701 |
| PH - Validation | 298 | 0 | 0 | 298 |
| PH - Test | 312 | 198 | 0 | 510 |

## 3.2 Preprocessing

To enable consistent cross-dataset comparison, all echocardiographic images were standardised using a geometry-aware preprocessing pipeline. A fan-shaped region-of-interest (ROI) mask was first applied to remove non-ultrasound background and irrelevant regions. The masked region was then cropped to the ROI bounding box and resized to 224 × 224 pixels using aspect-ratio-preserving scaling with zero-padding. The same spatial transformations were applied to the images, LV masks, and ROI masks to maintain alignment. This preprocessing was applied consistently across segmentation, handcrafted feature extraction, and latent-representation analysis. Full implementation details are provided in Appendix A. Implementation Details, while preprocessing ablation results are reported in Appendix E. Segmentation Preprocessing Ablation.

## 3.3 Handcrafted Image Features

A detailed description of the handcrafted feature extraction process and the computed features is provided in Appendix B. Feature Definitions. The corresponding code for this stage and the subsequent domain-transfer analysis is available in the project GitHub repository[1].

**Handcrafted Feature Configurations and Predictor Construction**

Feature extraction was performed under six controlled configurations, combining two intensity-processing conditions with three region-selection strategies, as shown in Fig. 3. Intensity normalization used z-scores. Features were extracted from whole-image, fan-shaped, LV, and near-LV regions, allowing the analysis to separate global appearance, fan-field context, and LV-related geometric effects.

[1] https://github.com/soroushelyasi/domain_shift

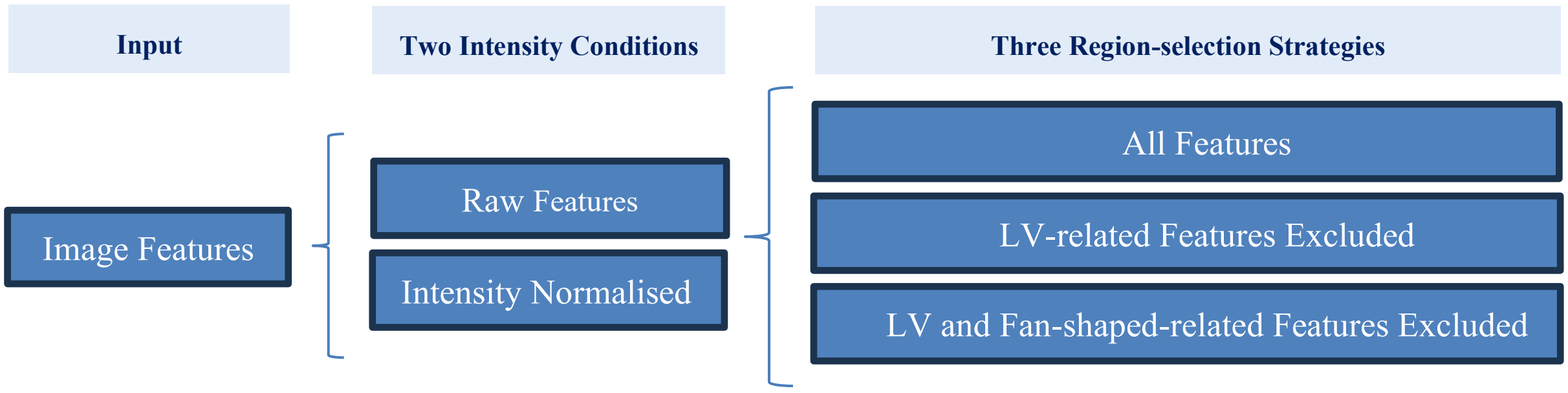

***Fig. 3. Six handcrafted-feature extraction configurations***

Numeric features were cleaned using missingness, variance, and correlation filters, then assigned to predefined interpretable feature families, covering intensity/contrast, image quality, texture/radiomics, edge and frequency descriptors, geometry/framing, and LV/near-LV boundary context. For domain-separability analysis, features were ranked using univariate statistical measures and classifier-based importance. For transfer-performance modelling, handcrafted features were transformed into source-normalised discrepancy predictors, including z-distance, signed z-distance, log-z-distance, and family-level distance summaries. Individual feature-distance predictors were selected within the training folds using mRMR, with subset size chosen through a sensitivity sweep to avoid information leakage. Full feature definitions, feature-cleaning thresholds, predictor-construction steps, and selection details are provided in Appendix A.6 Feature Cleaning, Selection, and Predictor Construction Details and Appendix B. Feature Definitions.

**Domain Separability Analysis and Visualisation**

Domain separability was evaluated to determine whether the handcrafted features contained dataset-specific signatures. First, univariate domain-separation scores were calculated for each feature using mutual information, ANOVA F-statistics, and maximum absolute Spearman correlation against one-vs-rest dataset labels. These scores were ranked and combined into an average univariate domain-separation score.

Supervised domain-classification models were also trained to predict the dataset label from the handcrafted feature representation. The implemented classifiers were balanced logistic regression, random forest, gradient boosting, and RBF-kernel support vector machine. Classification performance was evaluated using five-fold stratified cross-validation, with balanced accuracy and macro-F1 as the main metrics. Feature importance was estimated from classifier-native importances or coefficients and, when enabled, permutation importance using balanced accuracy as the scoring function. These outputs were combined to produce a final domain-separation ranking for each feature configuration.

Multivariate structure in the handcrafted feature space was examined to assess whether datasets formed distinct clusters after feature cleaning and standardisation. PCA was used to visualise dominant linear variation, while t-SNE and UMAP were used to explore non-linear dataset structure across the six feature configurations. Projection quality and dataset separation were summarised using trustworthiness, silhouette scores, clustering agreement, and centroid-distance diagnostics. These analyses provided complementary evidence of dataset separability beyond supervised domain classification. Full projection, clustering, and diagnostic settings are provided in Appendix A.7 Projection and Clustering Diagnostics.

**Transfer Performance Modelling and Validation**

Transfer performance modelling was used to estimate cross-dataset LV segmentation degradation from source-target discrepancy information. Same-domain transfers were excluded, and Dice drop was defined as the difference between the source model's mean within-domain Dice and its Dice score on each target image. Predictors broadly represented source-model characteristics, source-target image-domain discrepancies, and multivariate measures of domain similarity. Where individual feature-level discrepancies were included, predictors were selected within each training fold using an mRMR relevance-redundancy criterion to prevent information from the held-out transfers influencing feature selection. Detailed definitions and construction of the predictor sets are provided in Appendix A.8 Transfer Performance Modelling.

The regression analysis compared regularised linear models with non-linear ensemble models. Lasso, Ridge, and Elastic Net regression were used to support coefficient-based interpretation of standardised domain-shift predictors, while Random Forest and Histogram Gradient Boosting regression were included to capture potential non-linear relationships between image-domain discrepancy and segmentation transfer degradation. This model set was selected to balance interpretability and predictive flexibility.

Model validation used three complementary held-out schemes to assess generalisation under different transfer conditions. Leave-one-target-dataset-out (LOTO) excluded all transfers to a given target dataset, testing generalisation to unseen targets. Leave-one-source-dataset-out (LOSO) excluded all transfers originating from a given source dataset, assessing generalisation to unseen sources. Leave-one-source-target-pair-out (LOSTPO) excluded individual source-target pairs, evaluating unseen transfer relationships while allowing the corresponding datasets to appear in other training pairs. Together, these schemes assessed the robustness of transfer prediction across different levels of domain generalisation. Model performance was assessed using $R^2$, mean absolute error, root mean squared error, Pearson correlation, and Spearman correlation between the observed and predicted Dice-drop values. Full details of the predictor sets, mRMR-based feature selection, K sensitivity analysis, validation schemes, bootstrap confidence intervals, and evaluation metrics are provided in Appendix A.8 Transfer Performance Modelling.

**Feature Interpretation and Diagnostic Analysis**

Model interpretation and diagnostic analyses were conducted to identify the source-target discrepancies most strongly associated with Dice drop and to evaluate the robustness of the resulting predictors. These analyses comprised three complementary components, including association analysis, model-based interpretation, and robustness diagnostics.

Association analysis used Pearson and Spearman correlations to examine linear and monotonic relationships between discrepancy predictors and transfer degradation. Model-based interpretation used standardised Lasso coefficients and permutation importance to identify influential predictors and assess the direction and relative strength of their associations with Dice drop. Robustness was examined through block-ablation analysis, bootstrap confidence intervals, target-specific importance estimates, collinearity assessment, and scanner-vendor diagnostics.

Subsequently, evidence from domain-separation analysis, correlation with Dice drop, and transfer-model importance was integrated into a unified feature ranking. Features supported across multiple analyses were prioritised, providing a broader indication of the image-domain discrepancies most relevant to cross-dataset segmentation performance. Further methodological details are provided in Appendix A.9 Feature Interpretation and Diagnostics.

## 3.4 Variational Autoencoder (VAE) Features

In addition to handcrafted feature extraction, which provides high interpretability, we employed a VAE. A VAE provides a probabilistic framework that encodes images into a structured latent space and reconstructs them through a decoder network. This self-supervised objective enables the model to learn latent representations of intrinsic visual patterns, including intensity distributions, texture characteristics, and structural variations. These learned representations have been shown to correlate more closely with human perceptual judgements than low-level similarity measures [3], [26]. Furthermore, this approach does not rely on manually designed features, which may introduce bias through feature selection.

VAE-derived features were used to investigate domain shift across datasets acquired under varying imaging conditions. As with other DL models, the quality of the learned latent space depends on several design choices, including latent dimensionality and the balance between reconstruction and regularisation terms. Model selection and early stopping were based on validation reconstruction loss, with further implementation details provided in Appendix C. Modelling Details.

## 3.5 LV Segmentation and Segmentation Latent

In this study, LV segmentation was used to support analysis of domain characteristics across datasets. The resulting segmentation masks identify anatomically relevant regions, whilst features extracted from the segmentation network were used as task-specific latent representations. These representations provide a complementary perspective to the image-based latent features learned by the VAE. Unlike VAE features, which primarily capture global appearance characteristics, segmentation-derived features are expected to be closely related to anatomical structure and boundary information.

For this purpose, a U-Net-based architecture (Appendix C. Modelling Details) was trained independently on each of the five source datasets using their respective training and validation partitions. Each model was evaluated on its corresponding in-domain and external test sets without target-domain fine-tuning. The network followed a standard encoder-decoder design with skip connections, enabling the extraction of hierarchical features whilst preserving spatial detail. In addition to generating segmentation masks, bottleneck features were extracted and used as task-specific latent representations for domain shift analysis [31].

### 3.6 MMD and CMD Discrepancy Measures

Pairwise discrepancies between datasets were computed for handcrafted features, VAE latent features, and segmentation-derived latent features. For the handcrafted analysis, the six feature configurations described previously were used as input feature spaces. For each representation, samples were grouped by dataset label, and only datasets with at least two samples were retained for pairwise comparison.

To quantify domain shift between datasets, two distribution-based metrics, MMD and CMD, were employed [15], [20]. Latent vectors were obtained for each image and aggregated across the training, validation, and test splits to estimate a dataset-level distribution independent of the original data partitioning. Prior to discrepancy computation, all features were standardised once using global z-score normalisation across the pooled samples, rather than within the leave-one-pair-out regression folds. This step reduced scale-related differences and ensured that the discrepancy measures primarily reflected differences in distributional structure rather than absolute magnitude. The VAE was trained using the five source datasets together with RVENet. Its latent features were standardised across all seven dataset labels, including RVENet, while the pairwise discrepancy–Dice analysis was restricted to the six datasets included in the segmentation evaluation.

MMD was computed using a Gaussian radial basis function (RBF) kernel to quantify differences between dataset feature distributions. We used the biased empirical squared MMD ($MMD^2$), including the diagonal terms in the within-dataset kernel calculations. The kernel bandwidth was determined separately for each dataset pair using the median heuristic. Further implementation details are provided in Appendix A.11 MMD Implementation Details.

CMD was included as a complementary method to assess whether moment-based distribution matching can capture domain discrepancies as suggested by previous work in related contexts [21]. It was computed as the sum of Euclidean distances between corresponding moments, including the mean and higher-order central moments up to order five ($K = 5$), capturing differences in variance, skewness, and higher-order distributional structure.

For each dataset pair, both MMD and CMD were computed on the standardised feature representations, producing pairwise distance matrices for subsequent quantitative analysis and visualisation.

### 3.7 Histogram-Based Distribution Metrics and Wasserstein Distance

In addition to MMD and CMD, two histogram-based discrepancy measures were used to quantify distributional differences between datasets. These were Jensen-Shannon (JS) divergence and Kullback-Leibler (KL) divergence. These measures complement MMD and CMD by quantifying differences in empirical feature distributions, rather than comparing samples in the original feature space.

For JS and KL, image-level features from each representation were grouped by dataset label. Pairwise discrepancies were then computed between every pair of datasets. Because the feature spaces were multidimensional, histograms were estimated separately for each feature dimension using 64 equal-width bins. For each dataset pair, the bin range was determined by the joint minimum and maximum across both datasets and all feature dimensions, ensuring consistent bin edges within each comparison.

The marginal histograms were then averaged to obtain a compact dataset-level representation of the overall feature distribution, although this averaging did not preserve feature-specific differences.

Let P and Q represent the normalised histogram distributions of two datasets. JS divergence compares these distributions using a symmetric, smoothed form of KL divergence.

KL divergence is an asymmetric measure of information loss when one distribution is used to approximate another:

$$KL(P \parallel Q) = \sum_i P(i) \log \frac{P(i)}{Q(i)} \quad (1)$$

Because KL divergence is asymmetric, KL(P||Q) and KL(Q||P) generally produce different values. Therefore, both directional forms were computed, with symmetric summaries derived when required. JS divergence, a symmetric and smoothed variant of KL divergence, is defined as:

$$JS(P \parallel Q) = \frac{1}{2} KL(P \parallel M) + \frac{1}{2} KL(Q \parallel M) \quad (2)$$

where:

$$M = \frac{1}{2}(P + Q) \quad (3)$$

JS divergence is bounded and numerically stable, which makes it suitable for empirical histogram comparisons. A lower value indicates more similar dataset distributions, while a higher value indicates greater domain discrepancy.

Wasserstein distance, also known as Earth Mover's Distance, measures differences between probability distributions. In this study, W was defined as the average of the one-dimensional Wasserstein-1 distances across standardised feature dimensions. Each distance was calculated directly from the observed feature values without histogram binning, with equal weighting across dimensions, as shown in Eq. (4).

$$\mathrm{W}(A, B) = \frac{1}{d} \sum_{j=1}^{d} \int_{-\infty}^{\infty} \left| F_{A,j}(t) - F_{B,j}(t) \right| \mathrm{d}t \quad (4)$$

In Eq. (4), $F_{A,j}$ and $F_{B,j}$ represent the empirical cumulative distribution functions of standardised feature j in datasets A and B, respectively, and d is the number of feature dimensions. Each image contributed equally within its dataset. The Wasserstein-1 distance was calculated directly from the observed feature values without histogram binning and then averaged across all dimensions. Therefore, W represents the mean of one-dimensional Wasserstein-1 distances rather than a joint multivariate distance. It captures differences in individual feature distributions but does not account for dependencies between features. Raw W was used as the primary Wasserstein measure, while log(1 + W) was examined as a sensitivity analysis. Both raw and log1p-transformed $MMD^2$ and CMD were also evaluated.

All five discrepancy measures ($MMD^2$, CMD, JS, KL and W) were applied to the six handcrafted feature configurations, VAE latent features and LV-segmentation latent features. For each representation and discrepancy measure, the association between dataset discrepancy and pairwise Dice performance was evaluated using Pearson and Spearman correlation coefficients. The analysis included 15 unordered dataset pairs. Pairwise Dice was calculated as the mean of the two transfer directions where both were available, while pairs involving Unity Consensus used the single available direction. Simple linear regression was used to assess the relationship between discrepancy and Dice performance, with in-sample $R^2$ and mean absolute error (MAE) reported. Leave-one-pair-out cross-validation was also performed by fitting the regression on 14 pairs and evaluating it on the remaining pair, repeated across all 15 pairs. Cross-validated $R^2$ and MAE were calculated from the combined held-out predictions. Given the limited number of dataset pairs and their shared datasets, these analyses were considered exploratory.

To assess the sensitivity of the results to individual datasets, each of the six datasets was excluded in turn, removing its five associated pairs and leaving ten pairs for analysis. We also evaluated all 3,003 possible ten-pair subsets of the original 15 pairs to compare the dataset-exclusion results with subsets of equal size. For each subset, the regression models and leave-one-pair-out evaluations were repeated using the original discrepancy values and Dice outcomes, without repeating feature extraction or standardisation. The resulting percentile rankings were used to describe the sensitivity of the findings rather than to assess statistical significance.

# 4 Results

## 4.1 Cross-Dataset Segmentation Performance

Cross-dataset segmentation performance was evaluated by training on a single dataset and testing both in-domain and across external datasets. The U-Net architecture, incorporating the fan-shaped ROI masking and resizing strategy described in the methodology section, was used for this analysis. The corresponding Dice scores are presented in Table 3.

***Table 3. In-domain and cross-dataset LV segmentation Dice scores***

| Train | UN | CS | CA | ED | EP | HMC | UN:GE | UN:PH | CS:GE | CS:PH |
|---|---|---|---|---|---|---|---|---|---|---|
| UN | 0.9227 | 0.9333 | 0.8123 | 0.8327 | <u>0.7803</u> | 0.9094 | 0.9152 | 0.9236 | 0.9290 | 0.9334 |
| CA:GE | 0.6124 | 0.6068 | 0.9345 | <u>0.5142</u> | 0.6591 | 0.6958 | 0.5772 | 0.6055 | 0.7049 | 0.6042 |
| ED | 0.8277 | 0.8724 | <u>0.7890</u> | 0.9309 | 0.8673 | 0.8523 | 0.8181 | 0.8414 | 0.8829 | 0.8736 |
| EP | 0.7486 | 0.7734 | 0.7705 | 0.8757 | 0.8945 | 0.8275 | <u>0.7088</u> | 0.7587 | 0.7796 | 0.7739 |
| HMC | 0.8067 | 0.8011 | <u>0.6312</u> | 0.7767 | 0.7632 | 0.9263 | 0.7990 | 0.8094 | 0.7927 | 0.8028 |
| UN:GE | 0.8975 | 0.9164 | <u>0.7404</u> | 0.7982 | 0.7729 | 0.9115 | 0.9109 | 0.8936 | 0.9203 | 0.9161 |
| UN:PH | 0.8931 | 0.9364 | 0.7977 | 0.8355 | <u>0.7876</u> | 0.9106 | 0.8931 | 0.9222 | 0.9309 | 0.9367 |
| ALL | 0.9196 | 0.9372 | 0.9339 | 0.9300 | <u>0.8972</u> | 0.9036 | 0.9119 | 0.9218 | 0.9294 | 0.9393 |

**Note.** ALL denotes a single model trained on the pooled training sets of UN, CA, ED, EP, and HMC; CS was used only for evaluation because no training split was available. Underlined values indicate the lowest Dice score in each training row, identified from the reported Dice scores rather than a separately calculated transfer-degradation measure.

Geometry-aware preprocessing substantially improved several challenging cross-dataset transfers compared with the unprocessed setting (Appendix E. ). The largest improvement was observed when training on HMC-QU and testing on EchoNet-Dynamic, where Dice increased from 0.296 to 0.777. Similarly, HMC-QU to CAMUS improved from 0.243 to 0.631, while Unity to CAMUS increased from 0.599 to 0.812. These results demonstrate that preprocessing can substantially reduce transfer degradation, although improvements were not consistent across all dataset pairs. Transfer performance was also asymmetric, depending on the direction of transfer. For example, a CAMUS-trained model achieved a Dice of 0.514 on EchoNet-Dynamic, compared with 0.789 in the reverse direction, despite similar in-domain performance (0.935 and 0.931, respectively).

## 4.2 Dataset-Level Analysis

### Handcrafted Feature Analysis

All analyses were performed at the image level across six echocardiographic datasets. Before modelling transfer degradation, we quantified dataset separability in the handcrafted feature space using four classifiers across six feature configurations. As reported in Table 4, the datasets were highly separable, with dataset identity readily predictable from the extracted handcrafted features. High separability was maintained even after removing LV and fan-region descriptors and restricting the analysis to whole-image features.

The datasets therefore occupy systematically distinct regions of the handcrafted feature space, indicating substantial differences in their image characteristics. These differences are readily detectable by linear classifiers and remain evident after ROI masking and normalisation. Moreover, the minimal change after z-normalisation suggests that dataset separation is more strongly associated with texture, frequency, and morphological features than with brightness or contrast.

***Table 4. Dataset separability across feature configurations***

| Configuration | LR | RF | GBM | SVM-RBF | Features |
|---|---|---|---|---|---|
| All | **0.971** | 0.839 | 0.871 | **0.854** | 380 |
| All Normalised | 0.968 | 0.834 | 0.866 | 0.846 | 386 |
| Without LV/near-LV | 0.967 | **0.850** | **0.880** | 0.842 | 152 |
| Without LV/near-LV Normalised | 0.963 | 0.838 | 0.858 | 0.839 | 154 |
| Without LV and Fan-shaped | 0.954 | 0.840 | 0.857 | 0.840 | **87** |
| Without LV and Fan-shaped Normalised | 0.951 | 0.833 | 0.852 | 0.835 | 90 |

**Note.** Bold values indicate the best performance within each classifier column. Feature counts differ slightly because feature cleaning was applied independently to each configuration, resulting in different features passing the missingness, variance, and correlation filters.

The projections showed that the dataset structure was better captured by non-linear methods than by PCA, with UMAP generally producing the clearest separation. To examine this structure, PCA, t-SNE, and UMAP were applied to each feature configuration. As summarised in Table 5, t-SNE and UMAP preserved local structure better than PCA, while UMAP generally produced clearer dataset separation. The limited variance captured by the first two principal components indicates that the feature space is high-dimensional and poorly represented by a two-dimensional linear projection. The all-feature configuration produced the clearest separation under UMAP, whereas the best transfer-prediction performance was achieved using the normalised all-feature configuration.

The UMAP and t-SNE projections in Fig. 4 clarify relationships that classification accuracy alone cannot show. Consensus and Unity overlap in a shared cluster, suggesting similar engineered feature profiles. EchoNet Dynamic and EchoNet Pediatric form adjacent but separable clusters, consistent with related acquisition pipelines applied to different cohorts. HMC-QU appears as scattered fragments across the feature space. In contrast, CAMUS forms an isolated cluster in every projection, supporting its role as the most morphologically distinct cohort and the only held-out target where source-size, rather than image-level descriptors, dominated the permutation-importance ranking.

***Table 5. Projection quality across configurations and methods***

| Configuration | PCA Var (2D) | Trust. PCA/tSNE/UMAP | Silh. PCA/tSNE/UMAP | Features |
|---|---|---|---|---|
| All | 0.357 | 0.855 / 0.975 / 0.946 | **0.163** / **0.344** / **0.404** | 359 |
| All Normalised | 0.353 | 0.853 / 0.971 / 0.938 | 0.132 / 0.331 / 0.372 | 376 |
| Without LV/near-LV | 0.450 | 0.875 / **0.988** / **0.967** | 0.137 / 0.321 / 0.332 | 140 |
| Without LV/near-LV Normalised | **0.461** | **0.888** / **0.988** / **0.967** | 0.133 / 0.332 / 0.298 | 147 |
| Without LV and Fan-shaped | 0.443 | 0.857 / 0.987 / 0.965 | 0.077 / 0.323 / 0.348 | 82 |
| Without LV and Fan-shaped Normalised | 0.453 | 0.871 / **0.988** / **0.968** | 0.098 / 0.317 / 0.330 | 85 |

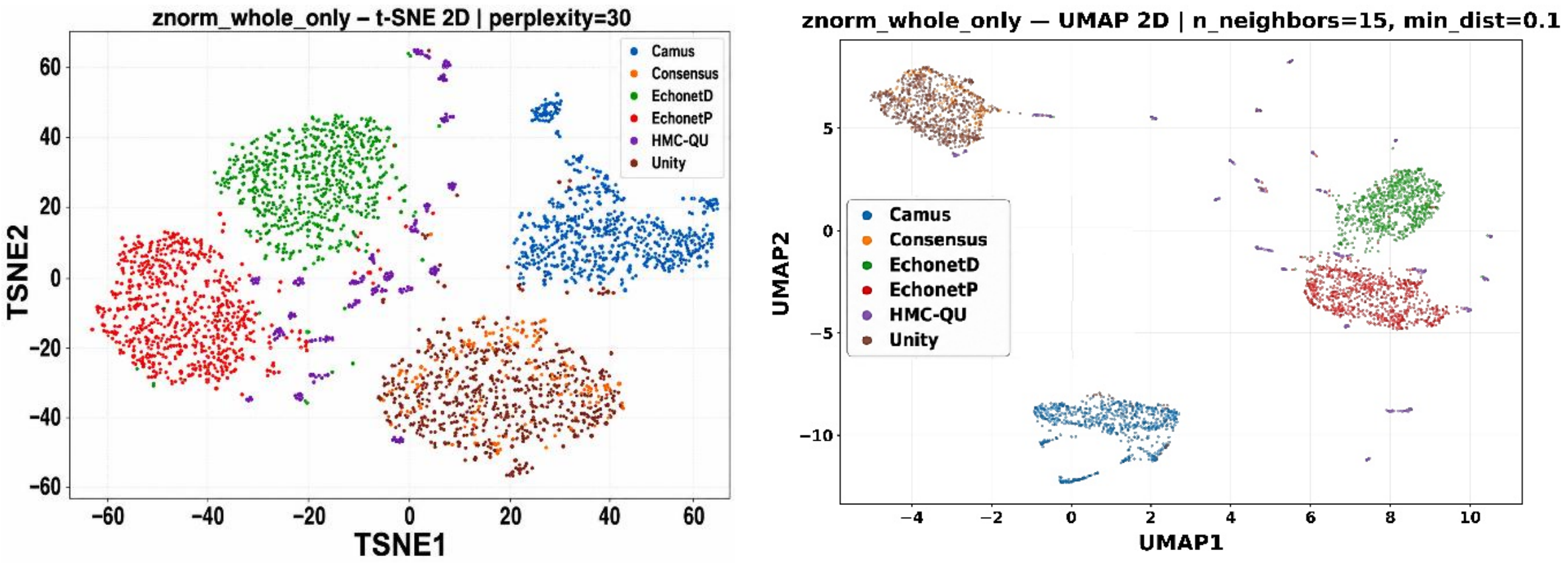

A B

***Fig. 4. Normalised 2D projections without LV and fan shaped features (A) t-SNE (B) UMAP***

Under held-out validation, segmentation degradation was moderately to strongly predictable across all six feature configurations (Fig. 5A). As shown in Table 6, under LOSTPO validation, the z-normalised full feature set achieved the best transfer-prediction performance, while the raw full-feature configuration performed similarly. Performance declined when LV/near-LV and fan-region descriptors were removed, indicating that these anatomical and contextual features contributed useful information for predicting Dice degradation.

Table 7 further shows that prediction remained moderately strong for unseen target datasets and unseen source-target pairs, but weakened when the entire source dataset was unseen.

***Table 6. Transfer prediction across feature configurations***

| Configuration | $R^2$ (Dice Drop) | 95% CI | Predictor Set | Best Model | Predictors |
|---|---|---|---|---|---|
| All | 0.602 | [0.589, 0.614] | Full image-domain | HGB | 123 |
| **All Normalised** | **0.612** | **[0.600, 0.624]** | **Full image-domain** | **HGB** | **143** |
| Without LV/near-LV | 0.492 | [0.480, 0.505] | Source-distance summary | RF | 12 |
| Without LV/near-LV Normalised | 0.500 | [0.487, 0.514] | mRMR z-family PCA/kNN | RF | 32 |
| Without LV and Fan-shaped | 0.418 | [0.408, 0.429] | mRMR log-z-distance | RF | 52 |
| Without LV and Fan-shaped Normalised | 0.476 | [0.467, 0.487] | mRMR z plus family | RF | 19 |

***Table 7. Transfer prediction across validation schemes***

| Config. | Best Config. | Best Model | Predictors | $R^2$ | MAE | RMSE | Pearson | Spearman |
|---|---|---|---|---|---|---|---|---|
| LOTO | All Normalised | RF | 73 | 0.609 | 0.0807 | 0.1126 | **0.783** | **0.712** |
| LOSO | All | Lasso | 53 | 0.476 | 0.0954 | 0.1304 | 0.696 | 0.680 |
| LOSTPO | All Normalised | HGB | 143 | **0.612** | **0.0820** | **0.1121** | 0.783 | 0.681 |

The comparison between raw and z-normalised configurations shows that intensity normalisation did not fundamentally change the ranking of feature sets (Fig. 5A). The full-feature and without LV/near-LV configurations changed only slightly after z-normalisation, while the whole-image-only setting improved more noticeably. This suggests that transfer degradation is not driven only by global brightness or contrast, but is also encoded in texture, frequency-domain, gradient-based, spatial-distance, and geometric descriptors that remain informative after intensity rescaling.

Fig. 5B shows predicted versus observed Dice drop for the best z-normalised all-feature model. The model captures a clear cross-target predictive signal, although the scatter shows dataset-specific heterogeneity and compression of the prediction range, suggesting that some CAMUS-specific shift is only partly captured by the engineered predictors. In contrast, Unity and many low-degradation cases cluster closer to the origin, indicating more benign transfer behaviour and lower segmentation degradation.

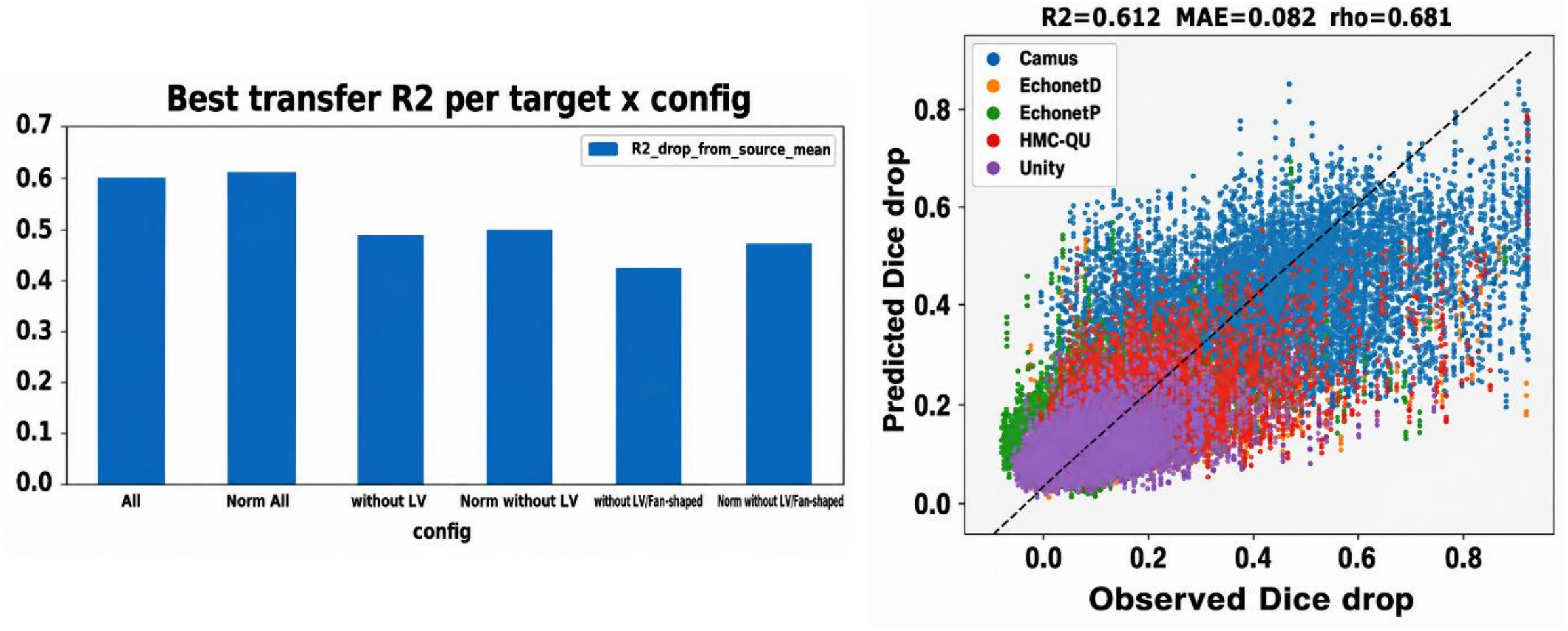

A B

***Fig. 5. Transfer-prediction performance across feature configurations. (A) Best $R^2$ for Dice drop (B) Observed versus predicted Dice drop for the best z-normalised all-feature model.***

Comparing the three anatomical-scope configurations showed that prediction performance declined as LV/near-LV and fan-region descriptors were removed. This indicates that anatomical and contextual features provided information beyond generic whole-image appearance and texture. However, within the datasets and transfer settings examined in this study, the model excluding LV/near-LV and fan-region descriptors still retained meaningful predictive performance. This suggests that whole-image and source-level features can be used to estimate transfer degradation without relying on LV-derived descriptors, although further validation on independent datasets is needed.

Bootstrap-stabilised Lasso coefficients were used to identify predictors that remained robust after multivariate regularisation, with Fig. 6 showing only descriptors whose 95% confidence intervals excluded zero. In the z-normalised all-feature setting (Fig. 6A), greater Dice drop was associated mainly with source-target distance, LV positional differences, geometry/framing, and selected LV/near-LV contextual characteristics. In contrast, image-quality, texture, edge-related, and source-level characteristics showed negative coefficients and were associated with smaller Dice drop in the multivariate model. When LV and fan-shaped descriptors were excluded (Fig. 6B), robust predictors shifted towards global image appearance, texture, source-distance, and source-level characteristics. Overall, the comparison shows that anatomical and contextual information contributes additional predictive signal, while global image characteristics remain informative when those descriptors are unavailable.

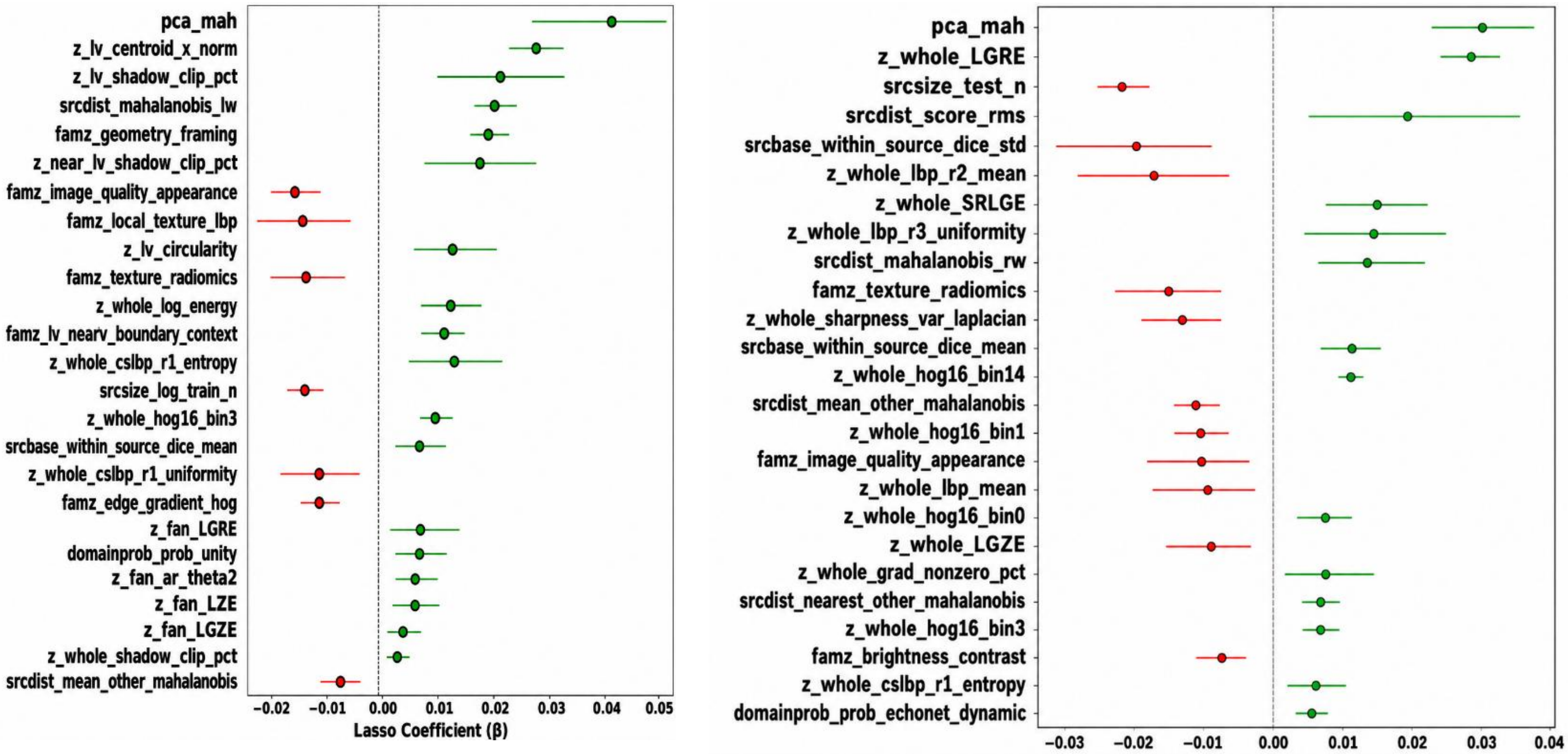

A B

***Fig. 6. Bootstrap-stabilised Lasso coefficients for Dice-drop prediction (A) normalised all features and (B) without LV and fan-shaped features.***

Fig. 7 summarises permutation importance for Dice-drop prediction. It indicates that the full model captures more anatomically interpretable factors associated with domain shift, whereas the reduced model relies more heavily on general image-texture and appearance statistics. In the z-normalised all-feature model (Fig. 7A), the largest reduction in predictive performance occurred when the kNN-based source-neighbourhood descriptor was shuffled. This was followed by LV centroid displacement features, particularly the normalised horizontal and vertical LV centroid deviations. These findings indicate that transfer degradation was strongly associated with a target image's distance from familiar source-domain examples and with differences in LV position or framing relative to the training distribution. Smaller contributions were observed for LV circularity, whole-image energy, LV high-frequency descriptors, radiomic texture features, brightness and contrast, and near-LV boundary-context features.

Fig. 7B also shows that, after removing the LV and fan-shaped descriptors, feature importance shifted towards global image appearance, texture, edge-related characteristics, and source-domain performance. In particular, the reduced model relied more strongly on whole-frame descriptors together with source-domain mean Dice. This suggests that, when local anatomical and fan-region information is unavailable, the model placed greater importance on global image characteristics and source-level performance information.

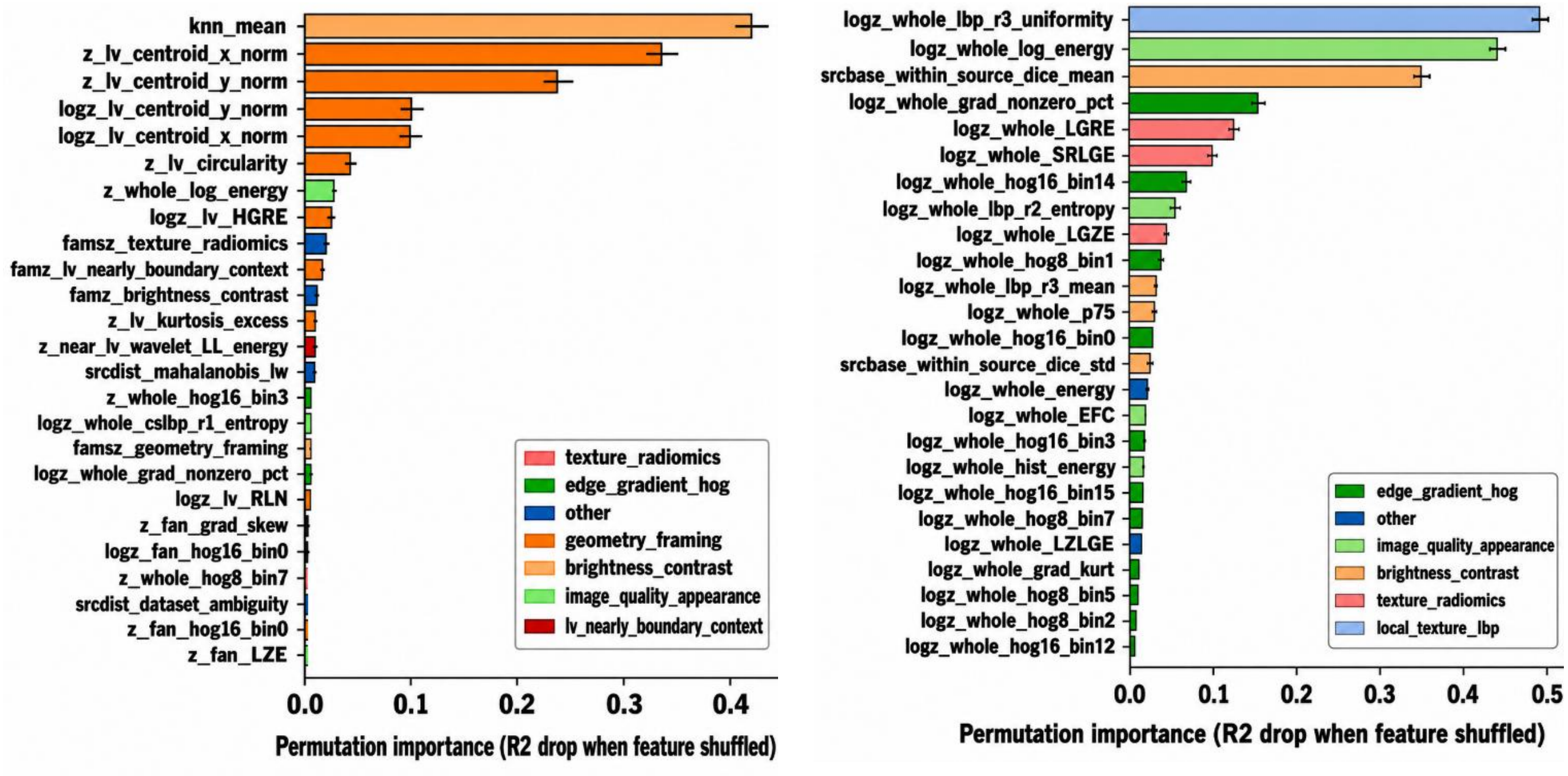

A B

***Fig. 7. Permutation importance for Dice-drop prediction using (A) z-normalised all features and (B) raw features without LV and fan-shaped descriptors***

Spearman analysis showed that source-target distance, edge/gradient, local texture, and image-quality descriptors were among the strongest associations with Dice drop across both feature configurations (Fig. 8A-Fig. 8B). In the z-normalised all-feature setting (Fig. 8A), additional positive associations were observed for source-baseline performance, image-to-source neighbourhood and PCA-based distances, geometry/framing, and texture/radiomics descriptors. When LV, near-LV, and fan-shaped information were excluded (Fig. 8B), source-distance and global whole-image characteristics remained strongly associated with transfer degradation. In contrast, source-size variables were generally negatively associated with Dice drop, indicating lower degradation for models trained on larger source datasets. Overall, the results show that transfer degradation is associated with a combination of source-target dissimilarity, image characteristics, anatomical context, and source-domain properties rather than any single descriptor. The feature-importance rankings from the best-performing model are provided in Appendix F. Feature-importance Analysis.

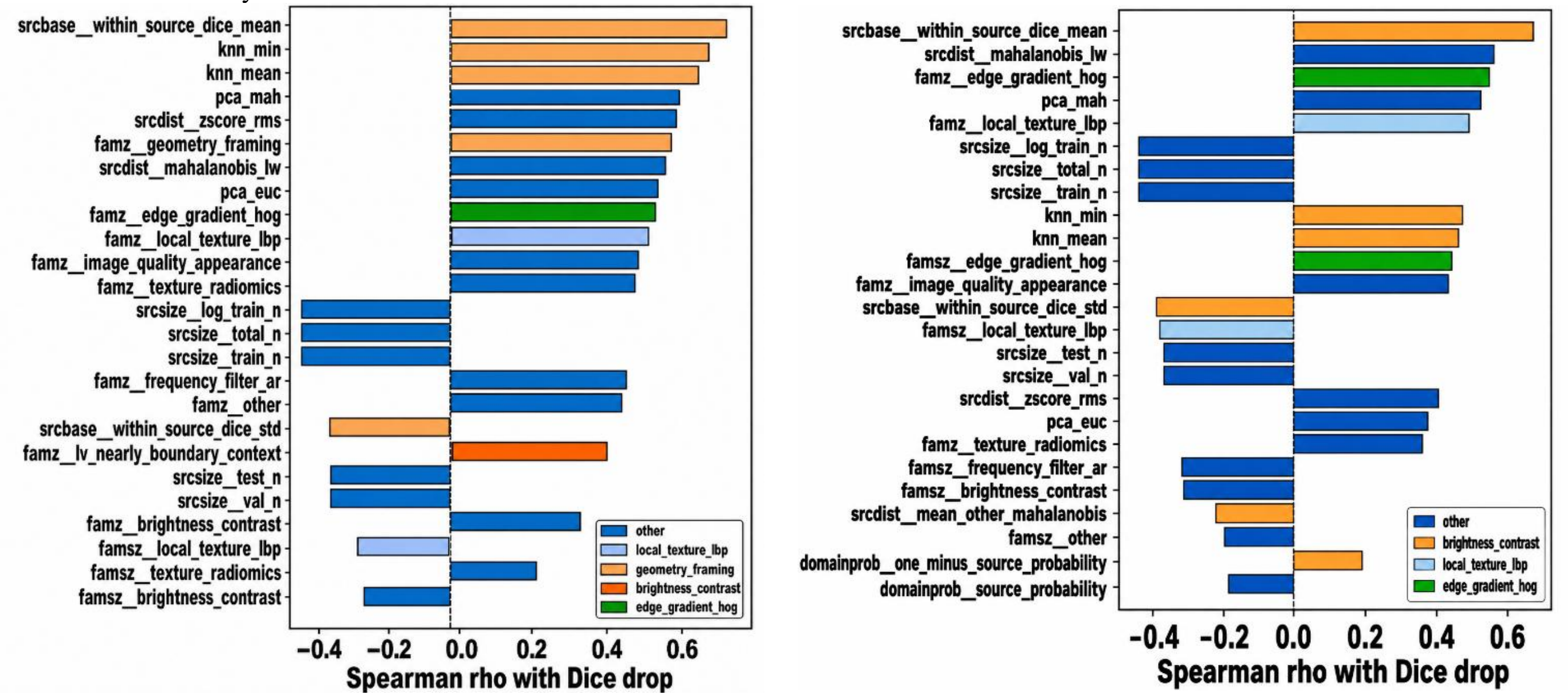

A B

*Fig. 8. Top 25 Spearman correlations with Dice drop for (A) z-normalised all features and (B) without LV and fan-shaped features; positive values indicate greater degradation.*

## Domain-Discrepancy Analysis

Domain discrepancy was evaluated by relating pairwise dataset-level distributional distances to pairwise Dice transfer performance. Five discrepancy measures were compared, including JS divergence, Wasserstein distance, KL divergence, $MMD^2$, and CMD. These were computed across handcrafted feature configurations covering B.2 BASIC INTENSITY STATISTICS to B.16 CROSS-REGION CONTRAST (LV vs NEAR-LV) as well as VAE and LV-segmentation latent representations. Fig. 9 summarises the correlations between discrepancy measures and pairwise Dice performance across the different feature representations, while Table 8 presents selected results for the main discrepancy measures.

A

B

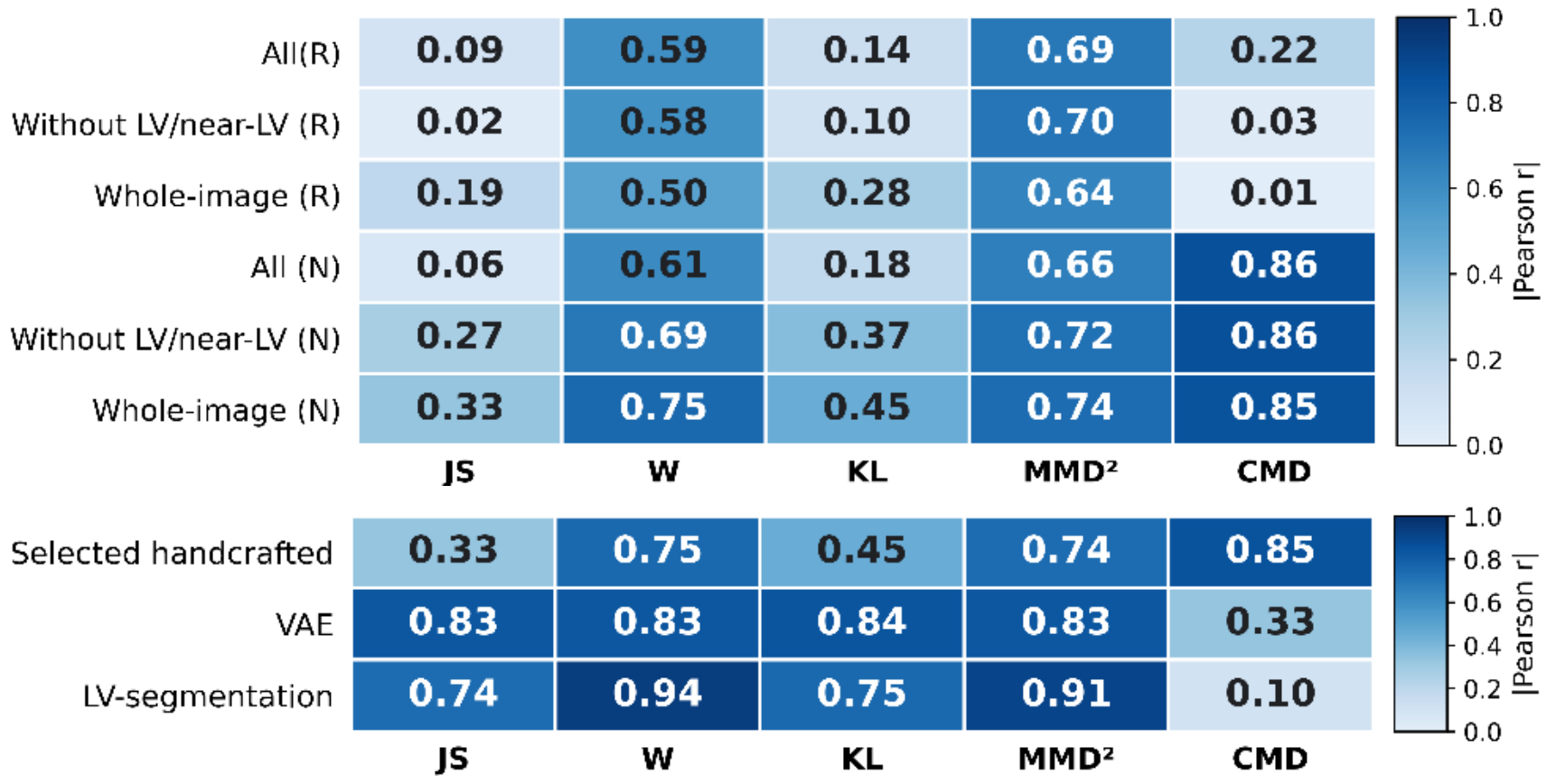

*Fig. 9. Absolute Pearson correlations between domain-discrepancy measures and pairwise Dice performance for (A) handcrafted feature configurations and (B) normalised whole-image handcrafted, VAE and LV-segmentation representations. R and N denote raw and normalised handcrafted configurations, respectively.*

The most informative discrepancy measure depended on both the representation and preprocessing strategy. MMD was generally strongest in the raw handcrafted feature spaces, whereas CMD became the leading handcrafted measure after z-normalisation. Importantly, strong associations were retained after excluding LV and fan-shaped descriptors, indicating that global image appearance and texture also capture transfer-relevant dataset differences. Both latent representations showed strong associations with pairwise Dice performance, particularly the LV-segmentation features using W. However, these associations should be interpreted as exploratory because they are based on only 15 dataset pairs, which are not independent as individual datasets appear in multiple comparisons.

The sensitivity analysis identified CAMUS as particularly influential. Excluding CAMUS substantially weakened the associations between dataset discrepancy and Dice performance for the latent and several normalised handcrafted configurations. Comparison with all 3,003 ten-pair subsets showed that this deterioration was unusual for these configurations, although the effect was not consistent across all discrepancy measures. Detailed results are provided in Appendix H. Supplementary Evaluation of Domain-Discrepancy Measures.

***Table 8. Selected discrepancy measures and descriptive pairwise Dice associations***

| Representation | discrepancy | Pearson r | $R^2$ | LOO $R^2$ | MAE |
|---|---|---|---|---|---|
| Normalised whole-image handcrafted | CMD | -0.852 | 0.726 | 0.640 | 0.040 |
| VAE latent | W | -0.830 | 0.689 | 0.590 | 0.042 |
| LV-segmentation latent | W | -0.940 | 0.884 | 0.852 | 0.027 |

## 4.3 Vendor-level Analysis

To investigate whether the degradation predictors identified at the dataset level remained valid at the vendor level, the dataset-level pipeline was repeated after regrouping the images so that each unique manufacturer-dataset combination was treated as a separate source rather than the dataset alone. The analysis included the three datasets for which per-image manufacturer metadata were available, namely CAMUS (GE), Consensus (GE and Philips), and Unity (GE and Philips). Vendor-level identification was less accurate than dataset-level identification, indicating greater overlap among vendor-defined groups, particularly within Unity, where GE and Philips samples showed substantial overlap in feature space. As shown in Table 9, transfer-prediction performance was also similar between the raw and z-normalised configurations, suggesting that brightness and contrast rescaling was not the main source of vendor-related shift. However, these findings should be interpreted cautiously because vendor metadata were available for only three datasets.

***Table 9. Vendor-level separability and transfer-prediction performance***

| Configuration | LR Bal-acc | $R^2$ (Dice) | Best Model |
|---|---|---|---|
| All | **0.902** | 0.571 [0.535, 0.605] | Lasso |
| All Normalised | 0.898 | **0.574** [0.536, 0.610] | Ridge |
| Without LV/near-LV | 0.890 | 0.510 [0.465, 0.550] | HGB |
| Without LV/near-LV Normalised | 0.851 | 0.523 [0.483, 0.556] | HGB |
| Without LV and Fan-shaped | 0.861 | 0.498 [0.452, 0.539] | RF |
| Without LV and Fan-shaped Normalised | 0.820 | 0.508 [0.471, 0.539] | Ridge |

Two-dimensional projections of the vendor-level source pool supported the separability results. Compared with the dataset-level analysis, vendor classes showed weaker local structure and limited separation in two dimensions. The small silhouette scores indicated that the five vendor groups did not form compact, clearly separated clusters, while the limited variance captured by PCA remained comparable to the dataset-level findings.

Fig. 10 shows that the main separation is still largely dataset-driven. When the same UMAP coordinates are coloured by dataset, CAMUS forms a distinct group, while Unity and Consensus form a separate cluster. When coloured by manufacturer label, GE does not form one coherent cluster. GE samples from CAMUS remain separate, while GE samples from Unity and Consensus are embedded within the Unity/Consensus cluster. Within that cluster, GE and Philips are mostly interspersed, with only a mild spatial gradient. The combined dataset-vendor analysis suggests that the observed vendor-level separation is largely influenced by the underlying dataset structure. In other words, GE CAMUS is distinct mainly because CAMUS is distinct, not because all GE images share one common manufacturer pattern.

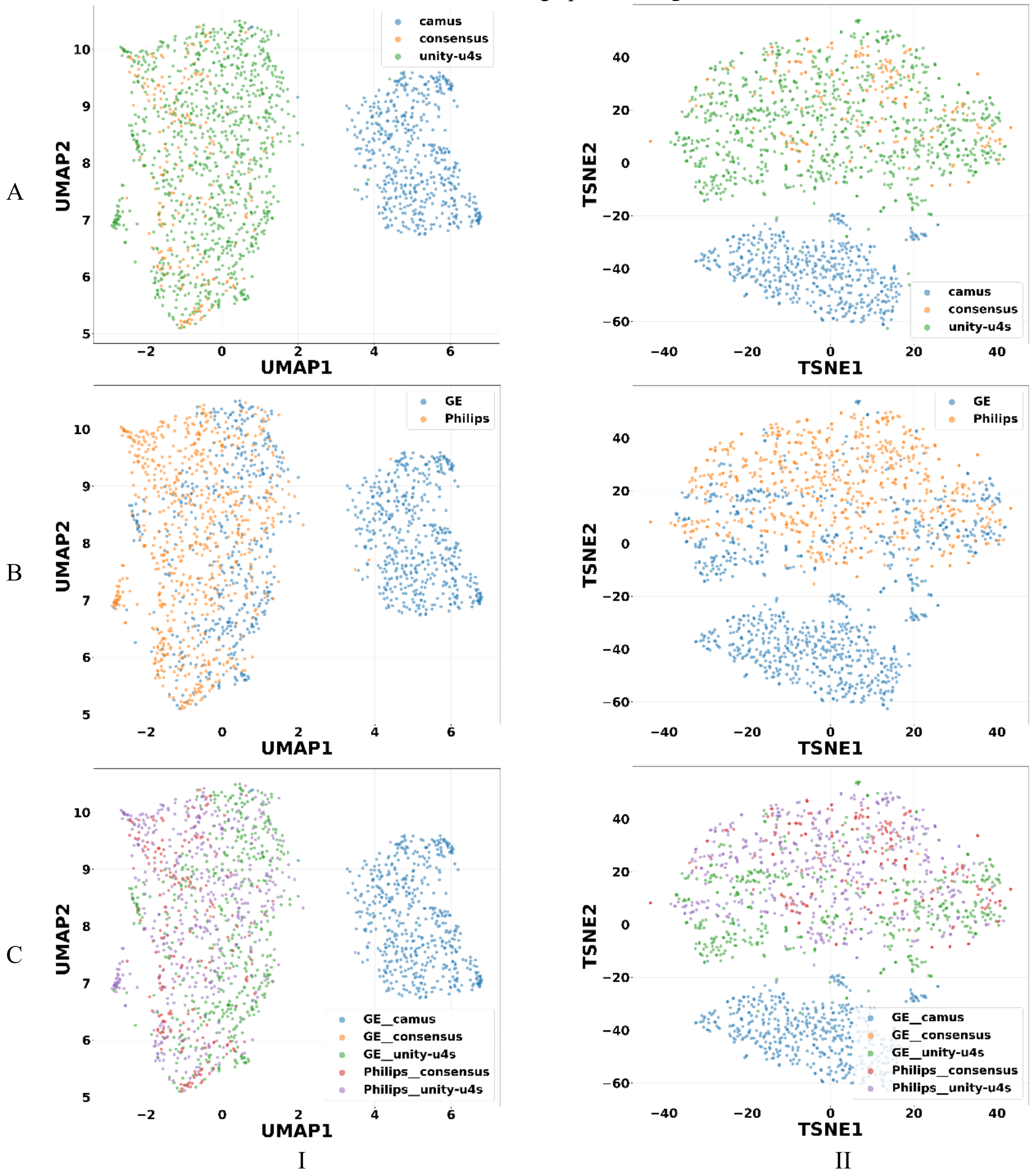

***Fig. 10. Vendor-level UMAP projections. Rows show (A) dataset labels, (B) vendor labels, and (C) dataset-vendor labels; columns show I) normalised all features II) without LV and fan-shaped features***

Lasso coefficients for the absolute Dice-drop target in the normalised all-feature configuration showed opposite-signed coefficients for the two PCA-based distance predictors, pca_mah and pca_euc, with standardised coefficients of +0.185 and -0.156, respectively (Fig. 11A). However, these predictors were highly collinear (Pearson $r = 0.996$), and both were positively associated with Dice drop in univariate analysis (Spearman $\rho = 0.73$). Furthermore, the pca_euc coefficient was non-zero in only three of the five held-out-target folds. These findings suggest that the opposing coefficients reflect instability associated with collinearity rather than independent effects in opposite directions.

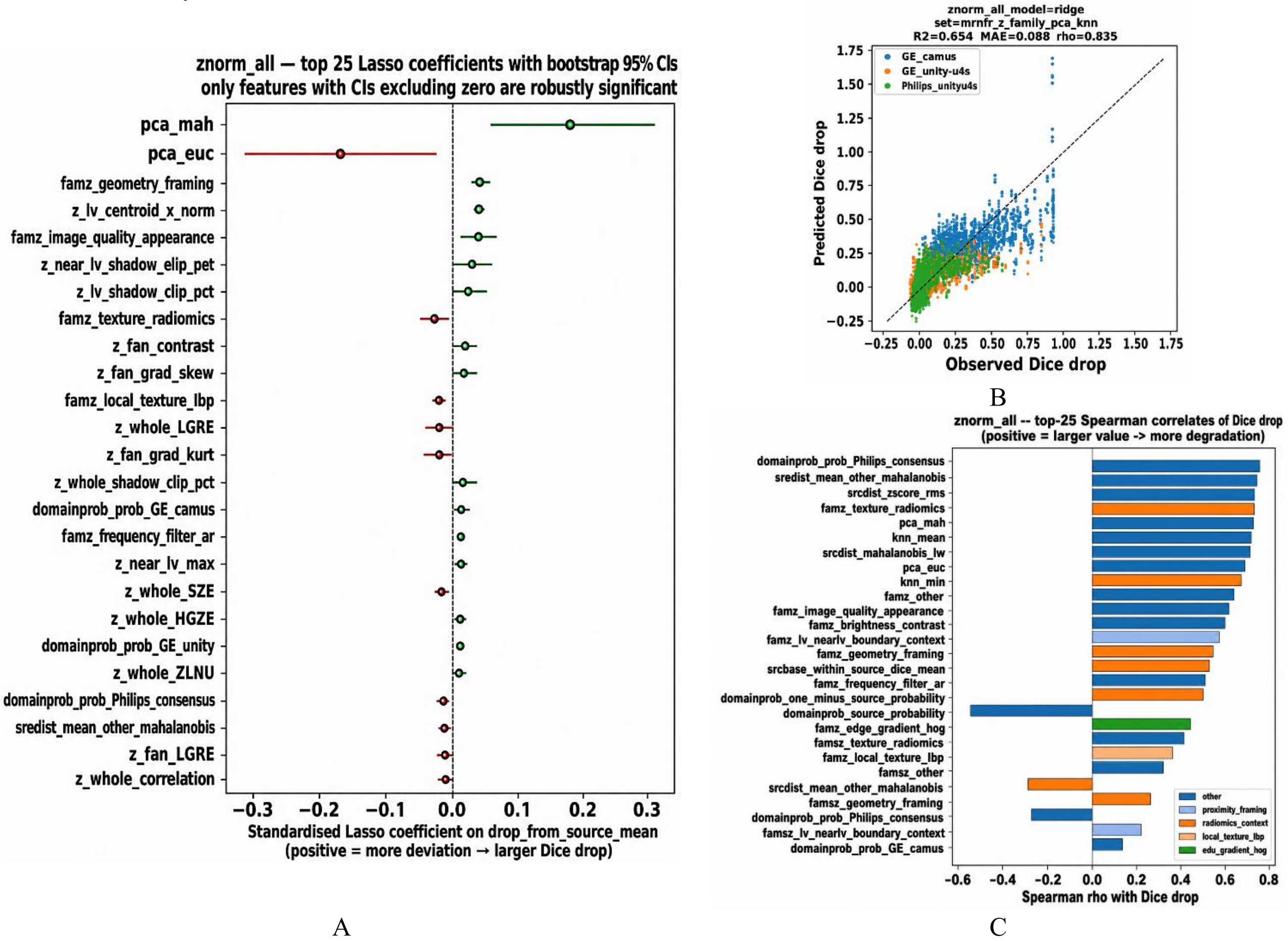

*Fig. 11. Predictors of Dice degradation (A) Robust Lasso predictors of Dice drop. (B) Observed versus predicted Dice drop from the ridge model. (C) Top Spearman correlations showing the strongest feature associations with segmentation degradation.*

Fig. 11B shows that predicted Dice drops broadly followed the observed values, particularly for smaller drops. Furthermore, Fig. 11C highlights associations with source performance, source-target distance, and image characteristics. Per-target permutation importance identified within-source Dice as the leading predictor in four of the five vendor groups, with GE CAMUS as the exception. Overall, these findings suggest that vendor-level transfer degradation is associated with source performance, domain similarity, and image geometry.

# 5 Discussion

## 5.1 Segmentation Transfer and Preprocessing Effects

The cross-dataset Dice scores in Table 3 empirically demonstrate transfer degradation in new environments. This degradation is asymmetric because CAMUS-trained models lose 0.23 to 0.42 Dice on other datasets, whereas models trained elsewhere lose only 0.11 to 0.30 when tested on CAMUS. This suggests that a source trained on a narrow appearance manifold transfers poorly to broader targets, while richer or more variable training data absorb a wider range of target characteristics. This pattern is consistent with the source-size and source-to-target distance effects identified later, indicating that training-data diversity is a structural property of the source rather than of any single transfer pair.

Vendor matching did not consistently improve cross-dataset transfer (Table 3). For example, although CAMUS is GE-based, transfer from CAMUS to Unity did not favour the GE subset, whereas the opposite pattern was observed for Consensus. Similarly, CAMUS transferred relatively well to EchoNet Pediatric despite substantial differences in patient age, while transfer to EchoNet Dynamic was poorer despite EchoNet Dynamic and EchoNet Pediatric originating from the same institution under similar acquisition settings. HMC-QU, although much smaller, also generalised relatively well in several settings. These findings indicate that dataset size, scanner vendor, and patient age alone do not consistently explain transfer performance. Scanner vendor therefore may not be the principal driver of domain shift in the datasets analysed. While scanner-related domain shift has been reported in MRI [24], the observed separation between echocardiography datasets sharing the same manufacturer suggests that other dataset-specific factors, such as acquisition protocols, institutional imaging practices, preprocessing and export settings, annotation conventions, and cohort composition, may also contribute to domain shift. However, apart from preprocessing, these factors were not individually evaluated in this study, and their relative contributions remain uncertain.

The data-split analysis (Table 2) also highlights a vendor imbalance in the Consensus test set, which was intended to support reliable evaluation of Unity. Therefore, the stronger performance on Consensus than on Unity may partly reflect this distributional bias toward Philips images rather than only more consistent segmentation labels, although a dedicated analysis would be needed to quantify this effect. More generally, many public datasets lack explicit scanner or vendor labels and combine imaging systems without distinction. Providing scanner type, vendor, and acquisition-protocol metadata in future releases would improve the reliability and interpretability of domain-shift analyses.

The segmentation experiments also clarify the role of preprocessing. Fan-shaped ROI masking, mask-aware cropping, and aspect-ratio-preserving resizing with padding improved several cross-dataset transfer cases compared with either no preprocessing or padding and resizing without prior cropping, as reported in Table 11 and Table 12 of Appendix E. . This improvement is likely due to preservation of the fan-shaped geometry and the spatial relationship between the near and far fields relative to the probe, allowing the network to learn LV position and shape more consistently across datasets. Thus, part of the apparent inter-dataset variability can be reduced through geometry-aware standardisation rather than model adaptation alone.

## 5.2 Dataset-level domain shift

### Interpretable Predictors (Handcrafted Features) of Transfer Degradation

Three principal findings emerge from the dataset-level handcrafted analysis. First, the six echocardiographic datasets are highly separable in the raw feature space. This is also supported by the image-level classification ablation in Appendix D. and challenges the assumption that a CNN trained on one cohort will generalise reliably to others without explicit adaptation. Second, segmentation degradation can still be predicted in advance. Together, these results show that dataset separability and transfer relevance are not equivalent. Consensus and Unity overlap in projection space, whereas CAMUS forms a separate cluster. However, transfer degradation is better explained by source-target distances across multiple feature families than by visual or classifier-based separability alone. Domain shift should therefore be summarised using several feature groups rather than a single separability score.

The third finding is the limited role of brightness and contrast, as z-normalisation had little effect on $R^2$ at both dataset and vendor levels. Instead, degradation is mainly encoded in texture, frequency-domain, gradient-based, and geometric descriptors that remain stable under global intensity scaling. These findings qualify the role of histogram matching, dynamic-range adjustment, and intensity normalisation in ultrasound harmonisation, which may address only part of the appearance variation captured by the classifier [32], [33]. The persistence of signal after z-normalisation is consistent with ultrasound texture studies showing that co-occurrence, run-length, gradient, frequency-domain, and other higher-order descriptors carry meaningful image information [16], [17], [34].

The projections in Fig. 4 add interpretation beyond classifier accuracy. Unity and Consensus overlap in the engineered feature space, EchoNet Dynamic and EchoNet Pediatric form adjacent clusters consistent with related acquisition pipelines applied to different cohorts, and HMC-QU appears fragmented, suggesting heterogeneous acquisition rather than one coherent imaging signature. CAMUS is isolated across all projections, identifying it as the most morphologically and statistically distinct cohort. This is consistent with its wider beam width and image-quality-stratified sampling [12], and helps explain why CAMUS-related transfer rows in Fig. 5B were systematically under-predicted. The concentration of the largest prediction errors in CAMUS-related transfers suggests that some CAMUS-specific aspects of domain shift are only partly captured by the engineered predictors.

Anatomical-scope analysis shows where the predictive information is located. In the raw-feature configurations, removing LV and near-LV descriptors reduced the Dice-drop $R^2$ from 0.602 to 0.492, while restricting the model to whole-image features without LV and fan-shaped ROI descriptors reduced $R^2$ to 0.418 using 52 predictors. This shows that transfer degradation can partly be predicted from region-agnostic image statistics. This is practically useful when segmentation outputs or anatomical priors are unavailable at inference time.

The recurrence of geometry and centroid features is also important. This suggests that some observed visual differences reflect variation in anatomical framing and acquisition context, where the LV is positioned, scaled, or oriented differently within the field of view. These effects may arise from acquisition practice, depth setting, probe position, field-of-view selection, and dataset-specific cropping or export conventions, so they are unlikely to be removed by intensity normalisation alone. This is consistent with previous MRI scanner-shift studies showing that acquisition parameters, protocols, and post-processing can influence image statistics and model generalisation [24], [25]. This may also help explain why the preprocessing pipeline helped increase Dice scores and supports the warning in Appendix E. Segmentation Preprocessing Ablation that high cross-dataset classification accuracy can partly reflect shortcut learning from framing or acquisition-context cues rather than genuine domain understanding.

The per-target analysis identified three broader factors associated with transfer degradation. First, source-model quality and training-data coverage were important, as models with stronger within-source performance and larger training sets generally transferred more reliably. CAMUS was the main exception. Second, degradation increased when target images were less similar to the source distribution, indicating that unfamiliar target samples were harder to segment reliably. Third, differences in anatomical framing, particularly LV position within the image, also contributed to transfer failure. These themes were consistent across permutation importance, Lasso coefficients, and correlation analyses, suggesting that transfer risk is shaped by source robustness, source-target similarity, and imaging geometry rather than by any single image descriptor.

At the vendor level, repeating the pipeline with source identity defined by manufacturer-dataset strata allowed scanner effects to be tested more directly. However, reliable image-level manufacturer metadata was available for only a limited number of datasets, so these results should be treated as exploratory. Even so, the evidence does not support manufacturer identity as the primary axis of domain shift.

Vendor-level classification showed lower separability than dataset-level classification. Moreover, the projection analysis in Fig. 10 supports this interpretation. CAMUS and the Unity-Consensus pool formed clearer groups when coloured by dataset than when coloured by manufacturer. GE images did not form a single cluster. GE-CAMUS remained distinct, whereas GE images from Unity and Consensus were embedded within the Unity-Consensus cluster alongside Philips images. This suggests that GE-CAMUS is distinct because of the characteristics of the CAMUS dataset, rather than because GE has a consistent manufacturer-specific signature. Apparent vendor effects therefore seem secondary to stronger dataset-level and institutional differences. Moreover, vendor-level transfer degradation remained predictable. Raw and normalised configurations differed only minimally, confirming that characteristics such as brightness and contrast scaling were not the main vendor-related sources of variation.

**Domain Discrepancy**

The discrepancy analysis in Table 8 and Fig. 9 shows that the usefulness of a domain-shift metric depends strongly on the representation in which it is applied. Metrics that performed well for handcrafted features were not necessarily the most informative in learned latent spaces. Domain shift should therefore not be characterised using a single metric independently of the underlying representation.

The comparison of representations also highlights a clear interpretive distinction. Task-specific segmentation features under raw W were most closely aligned with transfer performance consistent with their emphasis on anatomical and structural variations relevant to LV segmentation. VAE features captured broader image-appearance differences, including information useful for reconstruction but not necessarily important for segmentation. This distinction is consistent with previous evidence that deep feature representations capture perceptually meaningful image differences beyond conventional low-level similarity measures [26]. Handcrafted features were less task-specific but remained valuable because they are interpretable, clinically decomposable, and can reveal which appearance, texture, or geometric characteristics differ between datasets. Importantly, their association with transfer performance remained after excluding LV, near-LV, and fan-region descriptors, indicating that transfer-relevant differences can also be detected from global appearance and texture alone.

$MMD^2$ remained informative across handcrafted, unsupervised and task-specific feature spaces, consistent with earlier medical-image studies [20], [21]. However, in the full 15-pair analysis, W showed a stronger association with Dice performance than $MMD^2$ in the LV-segmentation latent space.

Fig. 19 and Fig. 20 support this interpretation. In Fig. 19, $MMD^2$ clearly separates CAMUS from the other datasets, consistent with its poorer generalisation to other datasets. Fig. 20 shows the same pattern quantitatively, with larger $MMD^2$ values associated with lower pairwise Dice in both learned representations. This relationship is stronger in the task-specific segmentation space, consistent with the interpretation that segmentation-derived features capture transfer-relevant differences. By contrast, CMD is less consistent across the latent representations, reinforcing that the effectiveness of a discrepancy metric depends on the feature space in which it is applied.

Handcrafted descriptors provide interpretability and can support mask-free assessment, VAE features enable analysis using unsupervised evaluation, and segmentation-derived features provide stronger task relevance when a trained segmentation model is available. Combining these perspectives may support future transfer-risk estimation and provide insight into the image characteristics underlying cross-dataset degradation.

**CAMUS as a Dataset-Specific Domain**

CAMUS was isolated from every other cohort in all projections and feature configurations (Fig. 4, Table 5), and it produced the most asymmetric transfer behaviour of any source (Section 5.1). Manufacturer identity does not account for this separation. GE-labelled images did not form a single cluster; GE and Philips samples from Unity and Consensus were interspersed, whereas GE-labelled CAMUS images remained apart (Fig. 10), and vendor-level separability was consistently lower than dataset-level separability (Table 4 and Table 9). Our analysis suggests that CAMUS behaves as a dataset-specific domain rather than simply as a GE-manufacturer subset. For qualitative context, representative GE-labelled images from CAMUS and comparator datasets are shown in Fig. 12.

Documented provenance provides some possible explanations. CAMUS was acquired at a single centre using one scanner model and one probe, with no protocol beyond clinical routine, and its sequences were exported from EchoPAC as polar B-mode data and resampled onto a grid defined in physical rather than pixel units [12]. Its image quality is graded and heterogeneous (35% good, 46% medium, 19% poor), half of the patients have an ejection fraction below 45% [12]. Although variations in image quality and vendor-related characteristics are expected to be represented in the latent feature space, CAMUS nevertheless forms a single, distinct cluster separated from the other datasets. This pattern suggests that dataset-specific acquisition, processing, and release protocols may have a stronger influence on its representation than image quality or vendor identity alone. CAMUS therefore combines slightly different acquisition and release provenance with unconstrained clinical imaging settings, which may contribute to its separation from the other datasets examined here.

Taken together, the preprocessing analyses help narrow the likely sources of the observed dataset separation. Z-normalisation left both dataset separability and transfer prediction essentially unchanged (Table 4 and Table 6), suggesting that simple global intensity differences, such as brightness and contrast, are unlikely to be the dominant factor. Geometry-aware preprocessing improved several of the poorer CAMUS transfer results (Appendix E. Segmentation Preprocessing Ablation), indicating that differences in field of view, anatomical positioning, and framing account for part of the transfer gap. At the same time, background and sector-boundary effects are less likely to fully explain the remaining separation because fan-ROI masking removes these regions consistently across all datasets. The residual dataset-specific structure therefore appears to reflect a combination of acquisition, processing, and cohort-related characteristics that is not removed by intensity normalisation or geometric standardisation alone.

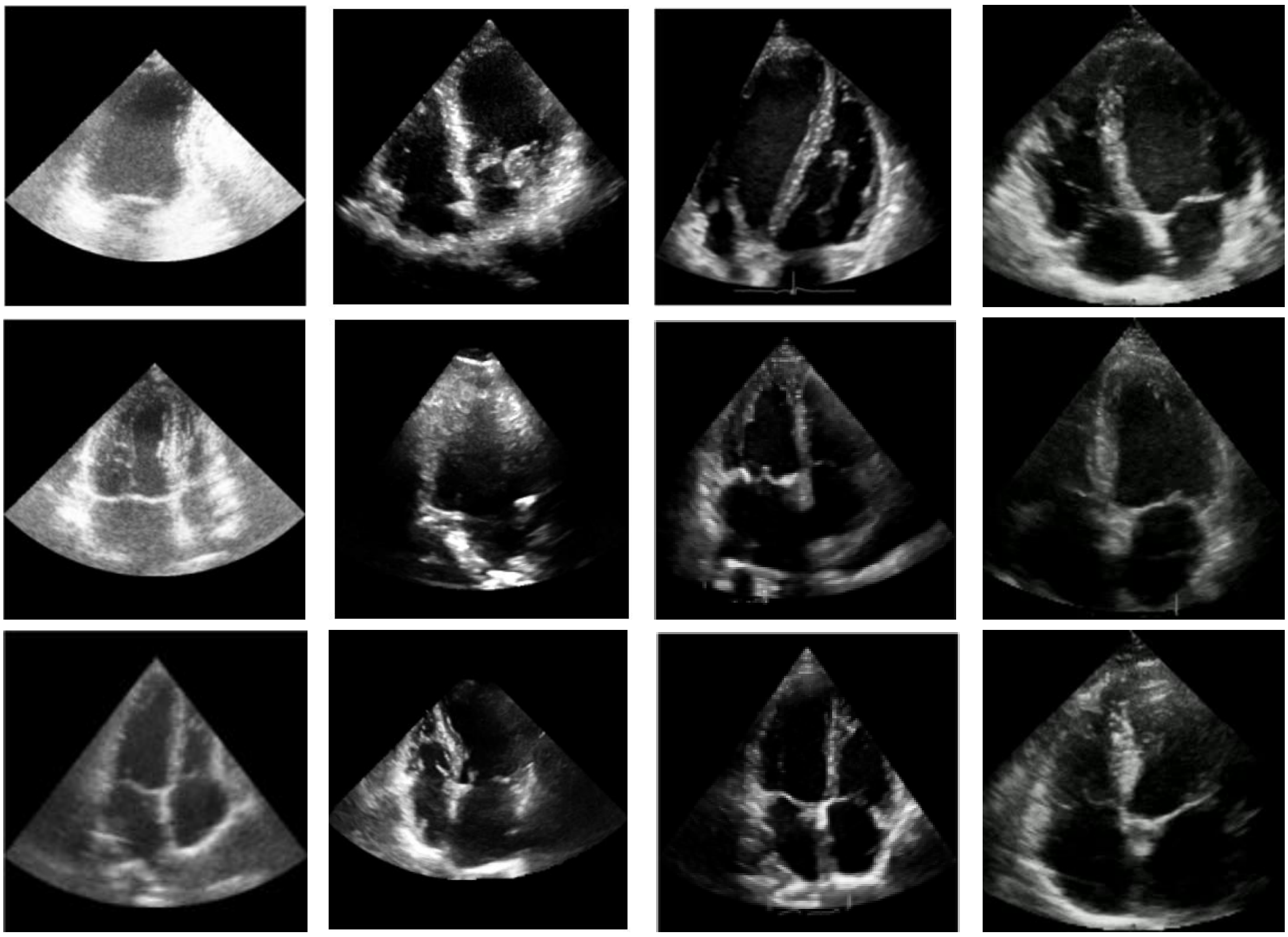

***Fig. 12. GE-labelled samples across different datasets [35], [36], [37]***

# 6 Conclusion

This study demonstrates that domain shift in echocardiographic LV segmentation is a structured and measurable phenomenon rather than an unpredictable consequence of external deployment. Across six datasets, geometry-aware preprocessing substantially improved cross-dataset segmentation, indicating that differences in field of view, anatomical framing, cropping, and resizing are an important component of apparent domain shift. However, the persistence of transfer degradation after preprocessing and intensity normalisation shows that higher-order texture, frequency, image quality, and geometric characteristics also contribute to limited generalisability.

Cross-dataset segmentation degradation could be anticipated using source-performance information and source-to-target feature discrepancies. Transfer risk was related to the robustness and coverage of the source dataset, the distance of target images from the source distribution, and differences in anatomical framing. Meaningful predictive information remained available after excluding LV-specific and fan-region descriptors, supporting the feasibility of transfer-risk assessment when segmentation masks or anatomical annotations are unavailable.

The comparison of handcrafted, unsupervised, and task-specific representations showed that no discrepancy measure was universally optimal. Handcrafted descriptors provided interpretable information about the image characteristics underlying domain shift, VAE representations enabled assessment using unlabelled target images, and segmentation-derived representations provided greater task relevance when a trained model was available. Although handcrafted features showed promising predictive value, their performance varied with the feature set and discrepancy measure, with the strongest association comparable to that of the VAE representation but weaker than that of the LV-segmentation representation. This suggests that manually designed descriptors may not fully capture the complexity of echocardiographic image formation and domain shift. Nevertheless, the complementary use of interpretable, unsupervised, and task-specific representations provides a practical framework for characterising and predicting cross-dataset transfer degradation.

The findings did not support scanner manufacturer as the principal determinant of domain shift in the datasets examined. Dataset identity, institutional acquisition practices, imaging protocols, export procedures, preprocessing, and cohort composition may contribute to transfer behaviour. These findings should, however, be interpreted in light of the limited number of datasets and dataset pairs. Because transfer was analysed at the dataset-pair level, the effective sample size for regression was small and covered a restricted range of acquisition settings. The observed relationships should therefore be regarded as indicative associations rather than causal evidence. Scanner metadata were also incomplete or unavailable for several publicly accessible datasets. Consequently, vendor-level analyses were restricted to datasets with identifiable manufacturer information, while the effects of scanner model, probe type, imaging settings, and other acquisition details could not be examined systematically.

Future studies should expand both data and metadata coverage by incorporating additional datasets, source-target pairs, vendors, institutions, and acquisition settings, alongside more comprehensive reporting of scanner models, probes, imaging protocols, and acquisition parameters. Further work should also explore additional image-derived descriptors to identify more informative and robust feature sets, integrate handcrafted and latent representations within unified modelling frameworks, and validate transfer-prediction models on larger independent dataset collections.

The proposed framework provides a basis for assessing segmentation generalisability before clinical deployment. With further validation, feature-distance measures could support source-dataset selection, identify target images likely to transfer poorly, and prioritise challenging cases for annotation, model adaptation, or additional data collection without requiring prior target-domain model training. This could reduce unexpected performance degradation and support more informed deployment of echocardiographic segmentation models in previously unseen clinical environments.

## Ethics

Ethical approval was granted by the University of West London Research Ethics Committee (UWL/REC/SCT-01753). The study used existing anonymised or de-identified echocardiographic datasets, with no participant recruitment or new data collection.

## Disclosure statement

The authors declare that there is no conflict of interest regarding the publication of this paper.

## Funding

This work was supported in part by the British Heart Foundation (grant no. RG/F/22/110059) and the Vice-Chancellor's Scholarship at the University of West London.

## Data availability

The Unity and Consensus datasets are not publicly available because of institutional data-use restrictions. However, access may be granted by the corresponding author upon reasonable request, subject to the required institutional approvals and data-use agreements. All other datasets are publicly available through their respective publishing institutions or official repositories.

# Appendix A. Implementation Details

## A.1 Fan-shaped ROI Masking

A fan-shaped region-of-interest (ROI) mask was applied to remove background pixels, annotation overlays, and non-ultrasound regions, ensuring that non-ultrasound background signals do not affect the analysis. The ROI masks were generated using a U-Net architecture with a ResNet-50 encoder pretrained on ImageNet, providing strong initialisation and improved generalisation. Input echocardiography images were converted to greyscale, normalised to [0,1], resized to 224×224, and replicated across three channels to match the encoder requirements. The trained network generated a binary segmentation corresponding to the fan-shaped ROI mask as shown in Fig. 13. The number of annotated ROI masks used for training is shown in Table 1, and the code for this stage is publicly available[2].

[2] https://github.com/soroushelyasi/echocardiography_fanshape_mask_extractor_new

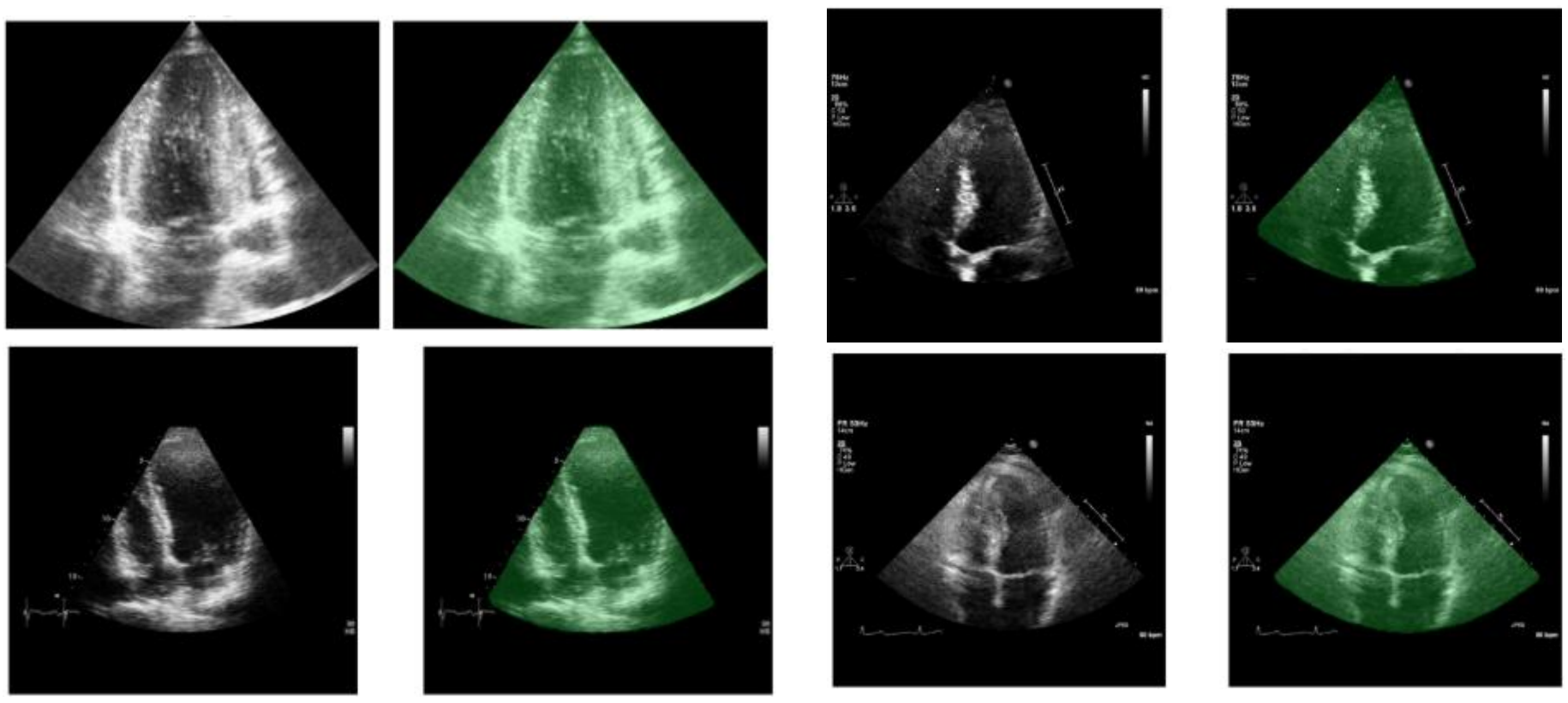

Original With Mask Original With Mask

*Fig. 13. Example fan-shaped ROI masks generated by ResUNet*

## A.2 Mask-Aware Cropping and Resizing

The fan-shaped ROI mask was first applied to remove irrelevant background regions. A bounding box derived from the ROI mask was then consistently applied to the image, LV mask, and ROI mask to preserve spatial alignment. The cropped region was resized to 224×224 using aspect-ratio-preserving scaling with zero-padding. This approach avoided geometric distortion whilst ensuring consistent spatial resolution across all samples, as illustrated in Fig. 14, where the preprocessing pipeline is applied to samples from Fig. 13. The ablation results showed that padding and resizing without mask-aware cropping did not consistently achieve the improvements obtained with the complete preprocessing pipeline (see Table 3 and Appendix E. ). Therefore, this preprocessing step reduced inter-dataset variability arising from differences in field of view and acquisition settings, supporting both image analysis and handcrafted feature extraction, and enabling more consistent anatomical representation across datasets.

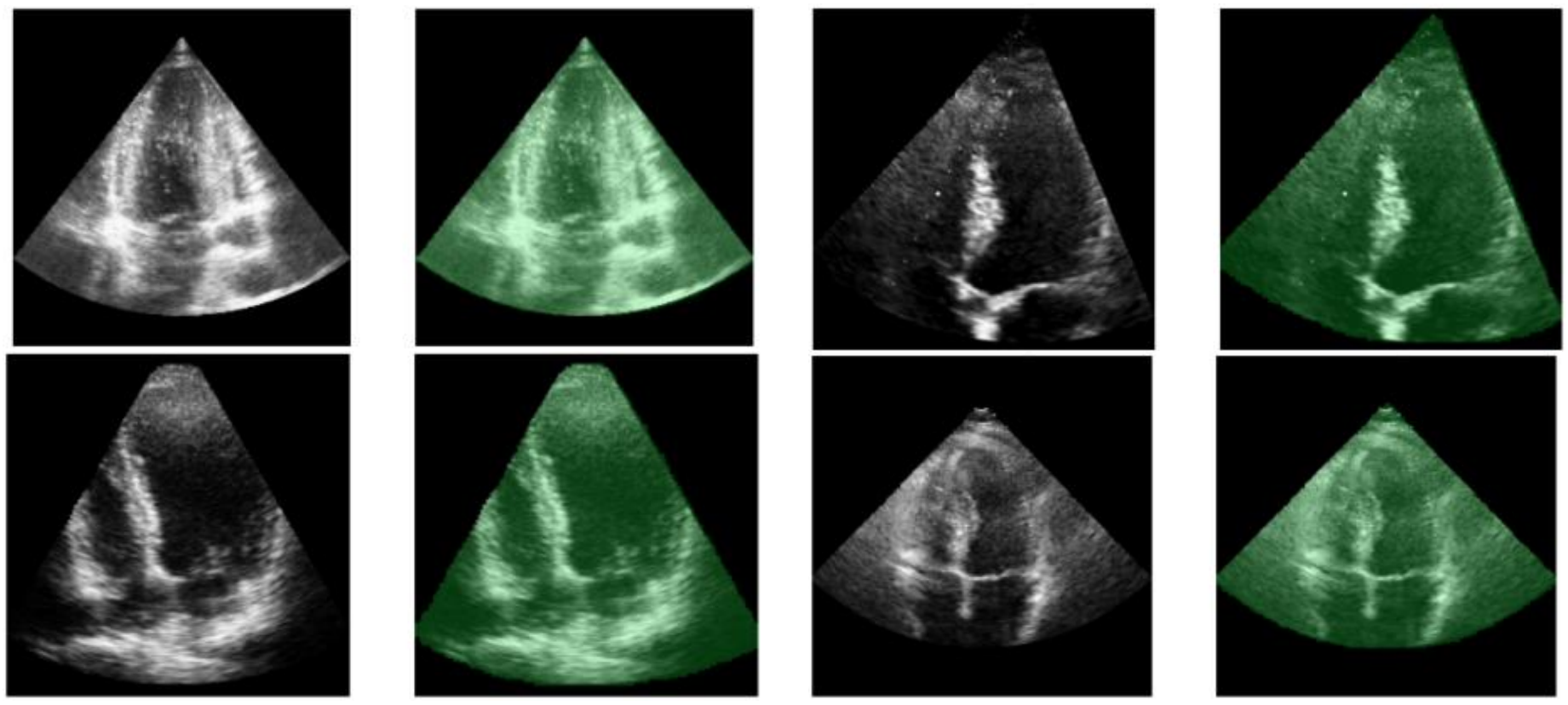

Original Preprocessed Original Preprocessed

*Fig. 14. Mask-aware cropping, padding, and resizing illustrations*

## A.3 ROI-Based Intensity Normalisation for Feature Extraction

For the handcrafted-feature analysis, features were extracted under both raw-intensity and ROI-normalised intensity conditions. After fan-mask-based cropping, padding, and resizing, raw features were first extracted from the preprocessed image. The image intensities were then z-score normalised using pixels inside the fan-shaped ultrasound ROI as the reference region. The resulting z-scores were clipped to the range [-3, 3] and linearly mapped back to an 8-bit range [0, 255]. The same feature-extraction procedure was then repeated on the normalised image.

## A.4 Fan-shaped Masking Network Training

The ROI masks were generated using a U-Net architecture with a ResNet-50 encoder pretrained on ImageNet. Training was performed using a batch size of 8 for a maximum of 200 epochs. The AdamW optimiser was used with a learning rate of $10^{-4}$ and weight decay of $10^{-4}$. To stabilise training and leverage transfer learning effectively, the encoder was frozen during the first 3 epochs and subsequently unfrozen for full fine-tuning. A ReduceLROnPlateau scheduler (factor = 0.5, patience = 2) was applied based on validation Dice score. Early stopping with a patience of 5 epochs was used to prevent overfitting. The loss function consisted of an equal-weight combination of Binary Cross-Entropy and Dice loss.

## A.5 Preprocessing Details

Following ROI masking, a geometry-aware preprocessing pipeline was applied. The fan-shaped ROI mask was first used to zero out all non-ultrasound regions. A bounding box was then extracted from the ROI mask and applied consistently to the image, LV mask, and ROI mask. To further refine the cropped region, a secondary tight crop was applied based on non-zero intensity regions within the masked image, removing residual empty areas. The resulting region was resized to 224×224 using aspect-ratio-preserving scaling, with zero-padding applied where necessary to maintain spatial consistency without distortion. Clustering based on the fan-shaped mask features is illustrated in Fig. 15 and Fig. 16.

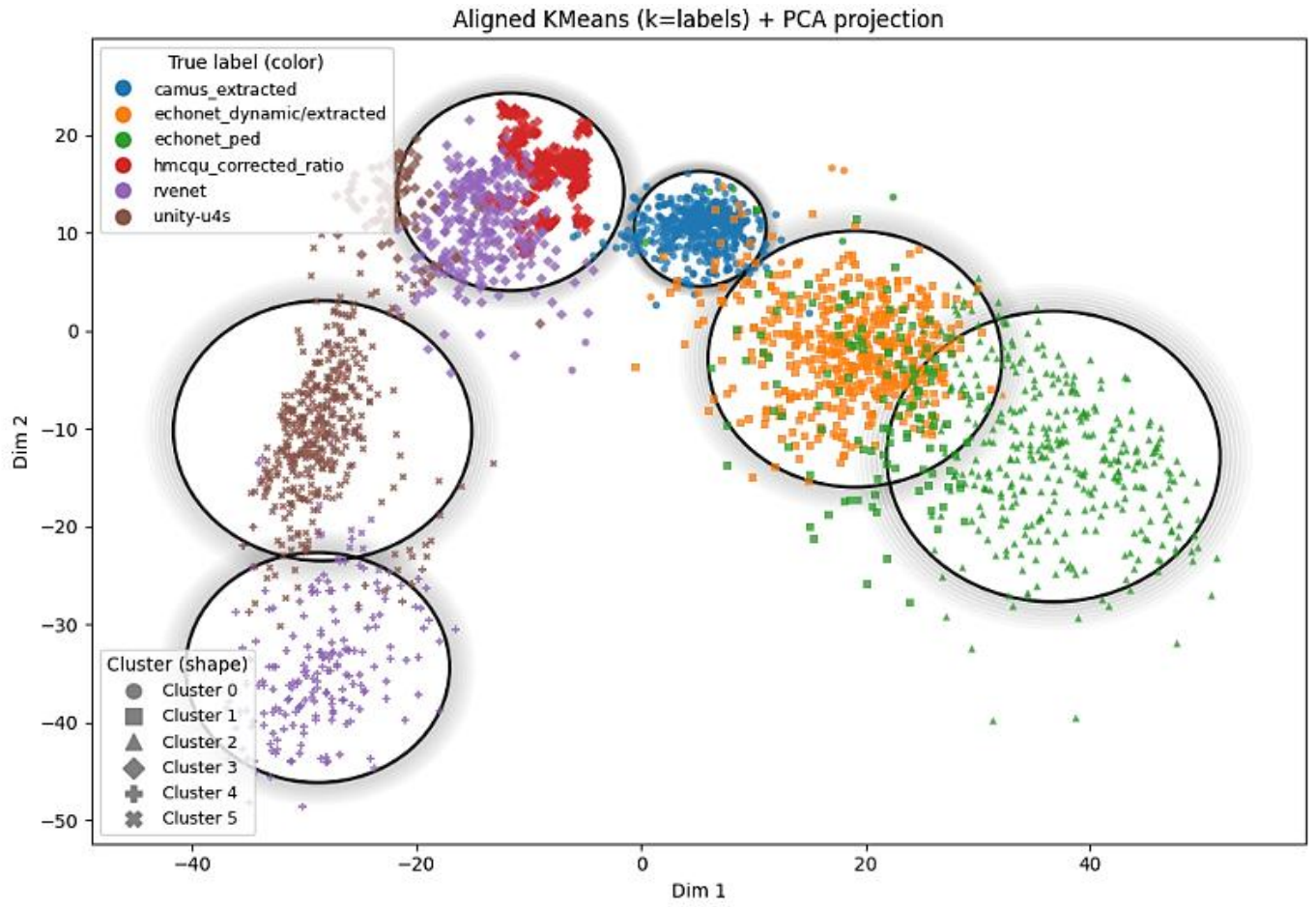

***Fig. 15. Dataset clustering based on fan-shaped ROI mask features.***

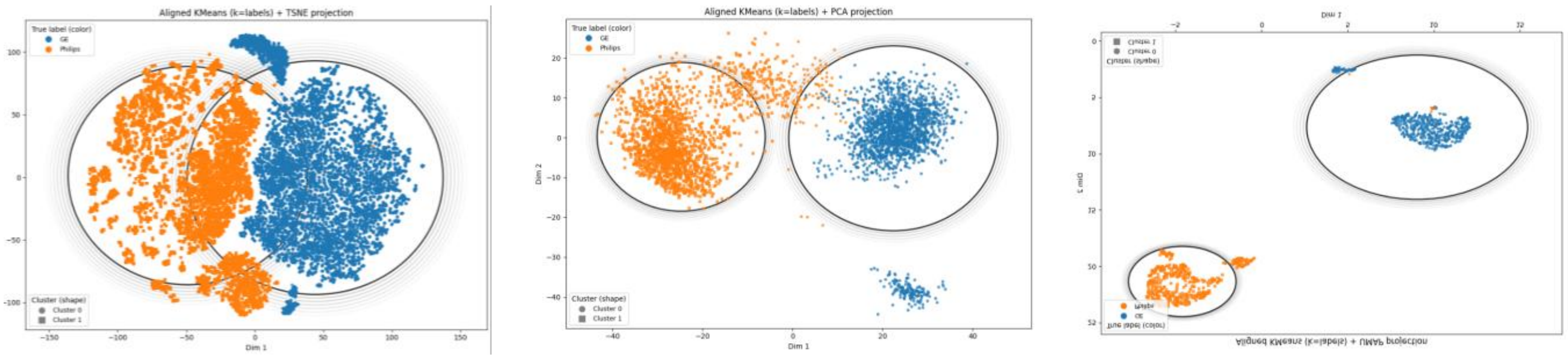

*Fig. 16. Vendor clustering based on fan-shaped ROI mask features (GE: blue, PH: orange).*

## A.6 Feature Cleaning, Selection, and Predictor Construction Details

For each feature configuration, metadata columns were separated from numeric feature columns. Numeric features were retained only if at least 50% of their values were non-missing and their variance exceeded the near-constant threshold of $1e^{-8}$. Remaining missing values were replaced using the feature median. For domain-separability and clustering analyses, a decorrelated feature matrix was also generated by removing highly correlated features above an absolute correlation threshold of 0.95.

Mutual information, ANOVA F-statistics, Spearman correlation, and classifier importance were used to rank features for dataset-domain separability. PCA was used for clustering diagnostics, source-target distance estimation, and projection visualisation. For transfer-performance modelling, z-distance, signed z-distance, log-z-distance, and family-aggregated distances were used as interpretable predictors. To construct compact predictor subsets, individual z-distance and log-z-distance features were selected inside each training fold using a minimum-redundancy maximum-relevance criterion, ensuring that feature ranking and redundancy filtering did not use held-out data. The number of retained features was selected using a K-sensitivity sweep, in which performance was profiled across candidate subset sizes and a stable operating point was selected for each configuration.

## A.7 Projection and Clustering Diagnostics

For each of the six feature configurations, numeric features were cleaned by removing mostly missing and near-constant variables, median-imputing remaining missing values, removing highly correlated features above an absolute correlation threshold of 0.95, and standardising the retained features. PCA was used to visualise the dominant linear directions of variation, while t-SNE and UMAP were used to visualise non-linear structure and potential dataset separation in two dimensions. For t-SNE and UMAP, PCA pre-reduction with up to 50 components was applied before projection to improve computational efficiency and reduce noise. Projection quality was summarised using trustworthiness and two-dimensional silhouette scores, and projection coordinates and scatter plots were saved for each feature configuration. Clustering diagnostics were also used to quantify dataset separability in the feature space. K-means and agglomerative clustering were applied using the number of known dataset classes as the number of clusters, and agreement between the resulting clusters and dataset labels was measured using the adjusted Rand index. Dataset centroids were computed in the standardised feature space, and pairwise Euclidean distances between centroids were used as an additional diagnostic of domain separation.

## A.8 Transfer Performance Modelling

Eleven predictor sets were evaluated to determine whether progressively richer domain-discrepancy representations improved the prediction of transfer degradation. Their composition and dimensionality are summarised in Table 10.

The original image features were not entered directly into the regression models. Instead, they were transformed into engineered source-target discrepancy predictors. For predictor sets containing individual feature distances, mRMR selection was performed separately within each training fold. Predictor ranking, redundancy filtering, and selection were performed using only the training portion of each fold.

Family-level summaries, PCA/kNN distances, domain-probability measures, source-level metadata, and extended source-distance descriptors were retained as aggregated predictors and were not subjected to individual mRMR selection. The largest predictor set therefore contained 143 engineered predictors rather than the full original image-feature space.

Under LOTO validation, all transfer rows associated with one target dataset were held out together. The model was trained using transfers involving the remaining target datasets, providing an evaluation of generalisation to an unseen target domain. Under LOSO validation, all transfers originating from one source dataset were excluded from training and used for evaluation, testing whether the learned predictor relationships remained stable for an unseen source domain. Under LOSTPO validation, complete source-target transfer pairs were held out, providing a transfer-level assessment in which the excluded relationship was unavailable during model fitting.

All predictor selection, K-sensitivity analysis, standardisation, and model fitting procedures were repeated independently within each validation fold. This nested procedure prevented information from the held-out transfers from influencing predictor construction or model optimisation.

The reported error metrics quantified complementary aspects of performance. $R^2$ measured explained variance, mean absolute error represented the average magnitude of prediction error, and root mean squared error placed greater weight on larger errors. Pearson correlation measured linear agreement, while Spearman correlation assessed agreement in the ranking of observed and predicted Dice-drop values.

For the best-performing model, 95% confidence intervals for $R^2$ were estimated using 1,000 bootstrap resamples. Standardised Lasso coefficients were expressed as the expected change in Dice drop associated with a one-standard-deviation increase in the corresponding predictor, allowing coefficient magnitudes to be compared across predictors measured on different scales.

*Table 10. Predictor sets for transfer-performance regression analysis*

| Predictor Set | Included Predictors | All Features | No LV-Near-LV | No LV-near-LV and Fan-shaped |
|---|---|---|---|---|
| Source baseline | Within-source Dice mean and standard deviation | 2 | 2 | 2 |
| Source baseline + size | Source baseline + source train/validation/test/total/log-size descriptors | 7 | 7 | 7 |
| Source distance summary | Source baseline + PCA/kNN distances + extended source-distance descriptors | 12 | 12 | 12 |
| Family distance | Family-level absolute z-distance + signed z-distance summaries + source baseline | 20 | 18 | 16 |
| mRMR z-distance | Training-fold selected absolute z-distance predictors + source baseline | K* + 2 | K + 2 | K + 2 |
| mRMR log-z-distance | Training-fold selected log-z-distance predictors + source baseline | K + 2 | K + 2 | K + 2 |
| mRMR z + family | Training-fold selected z-distance predictors + family absolute z-distance summaries + source baseline | K + 11 | K + 10 | K + 9 |
| mRMR log-z + family | Training-fold selected log-z predictors + family absolute z-distance summaries + source baseline | K + 11 | K + 10 | K + 9 |
| mRMR z + family + PCA/kNN | Training-fold selected z predictors + family z summaries + source baseline + PCA/kNN distances + domain-probability predictors | K + 23 | K + 22 | K + 21 |
| mRMR z + log-z + family | Training-fold selected z and log-z predictors + absolute and signed family summaries + source baseline | 2K **+ 20 | 2K + 18 | 2K + 16 |
| Full combined | Training-fold selected z and log-z predictors + family summaries + source baseline + PCA/kNN + domain-probability + source-size + source-distance descriptors | 2K + 43 | 2K + 41 | 2K + 39 |

** K = number of mRMR-selected individual z/log-z distance predictors; K was tested using sensitivity analysis over 10, 20, 30, 40, and 50, with selection performed inside each training fold.*

***mRMR selected K features from the z-distance pool; the log-z representations of the same K features were included alongside, yielding 2K candidate predictors per fold. Lasso regularisation resolved collinearity between each z and log-z pair.*

## A.9 Feature Interpretation and Diagnostics

Individual source-normalised z-distance features were ranked according to the magnitude of their Spearman correlation with Dice drop. Lasso coefficients were interpreted after predictor standardisation; positive coefficients indicated that greater discrepancy was associated with a larger Dice drop, whereas negative coefficients indicated the opposite relationship. Permutation importance was calculated using the complete modelling pipeline so that imputation and scaling were preserved during feature shuffling.

For block-ablation analysis, predictor groups were added progressively, beginning with source-quality predictors and followed by distance summaries, selected z-distance features, feature-family summaries, and domain-probability predictors. Lasso coefficient stability was assessed using bootstrap confidence intervals, and target-specific permutation importance was calculated by evaluating each held-out target dataset separately. Variance inflation factors were used to examine collinearity among source-quality predictors, while vendor diagnostics compared Dice-drop distributions according to whether a given scanner vendor was represented in the source dataset.

The unified ranking combined feature ranks from domain-separation analysis, z-distance correlation with Dice drop, and transfer-model permutation importance. The final score was calculated as the mean rank across the available evidence sources, with greater interpretive emphasis placed on features supported by multiple analyses.

### A.10 Implementation Environment

All experiments were implemented in PyTorch and trained using CUDA on a single NVIDIA RTX 4090 GPU (24 GB VRAM). The remaining components of the proposed framework were implemented as a Python-based pipeline, leveraging standard scientific computing and machine learning libraries including NumPy, SciPy, and scikit-learn. To ensure reproducibility, a deterministic random seed was used across all stochastic components of the analysis. The pipeline was designed with automated logging and output validation, enabling systematic tracking of intermediate steps, verification of generated results, and consistent organisation of outputs across experiments.

### A.11 MMD Implementation Details

For two datasets containing feature vectors X and Y, the biased empirical squared MMD was calculated using the within-dataset and between-dataset Gaussian kernel evaluations, as defined in the accompanying equation. The Gaussian RBF kernel was defined as $k(x, y) = \exp(-\gamma\|x - y\|^2)$. For each dataset pair, $\gamma = 1/(2D)$, where D is the median of the finite, positive, off-diagonal squared Euclidean distances among pooled feature vectors. If D was unavailable or no greater than $10^{-12}$, $\gamma$ was set to 1. The within-dataset sums included diagonal terms, and all samples were retained without subsampling.

## Appendix B. Feature Definitions

All original image-level features were computed across four regions, including whole_, fan_, lv_, and near_lv_. The base extractor covered the descriptors listed in B.1 METADATA to B.16 CROSS-REGION CONTRAST (LV vs NEAR-LV). During analysis, additional transfer predictors were generated from B.17 Transfer Metadata Columns onwards. In the table, 554 base feature variables were represented as z__, sz__, and logz__ distance forms, producing 1,662 feature-distance columns. With metadata, family summaries, PCA, kNN, source-distance, domain-probability, source-baseline, source-size, and target-outcome columns, the final table contained 1,718 columns.

The extracted features captured complementary properties of the imaging data. First-order intensity features described pixel-value distributions, including brightness, contrast, dynamic range, percentiles, entropy, histogram energy, clipping, sharpness, noise estimate, speckle contrast, and edge density. Texture was quantified using several descriptors, including GLCM, GLRLM, GLSZM, NGTDM, LBP, Gabor, wavelet, HOG, gradient, and autoregressive features (the full feature set is available in Appendix B. Feature Definitions). These features provided interpretable summaries of ultrasound appearance, speckle behaviour, local texture, directional structure, and image quality [15], [16], [17], [18], [19], [34].

### B.1 METADATA

- image_id: Unique identifier for the image

- split: Train/val/test assignment
- dataset: Source dataset name
- image_path: File path to the image

## B.2 BASIC INTENSITY STATISTICS

- valid_pixel_fraction: Proportion of pixels inside the region mask (useful for capturing differences in field of view, fan size, LV size, and valid ultrasound area)
- mean: Average pixel intensity (useful for capturing global brightness differences caused by gain, dynamic range, scanner settings, or post-processing)
- median: Middle pixel intensity value (useful for capturing typical brightness while being less affected by extreme bright or dark pixels than the mean)
- std: Standard deviation of intensities (useful for capturing contrast variability, speckle strength, and image heterogeneity)
- min/max: Lowest/highest pixel value in the region (useful for detecting intensity range differences, clipping, black-background dominance, or very bright structures)
- p1, p5, p10, p25, p75, p90, p95, p99: Percentile values of the intensity distribution (useful for describing the full brightness distribution and detecting differences in gain, compression, shadowing, and bright highlights)
- variance: Spread of intensities, equivalent to standard deviation squared. (useful for capturing overall image variability and contrast strength)
- cv: Coefficient of variation, calculated as standard deviation divided by mean. (useful for measuring relative variability independent of absolute brightness, which helps compare images with different gain levels)
- skewness: Asymmetry of the intensity histogram (useful for detecting whether an image is dominated by dark background, bright tissue, or asymmetric intensity distributions)
- kurtosis_excess: Peakedness of the histogram relative to a normal distribution (useful for identifying images with unusually concentrated or heavy-tailed intensity distributions, often linked to clipping or strong artefacts)
- entropy_bits: Shannon entropy of pixel intensities in bits (useful for measuring intensity complexity; low values may indicate uniform images, while high values may indicate richer texture or noise)
- hist_energy: Sum of squared histogram bin probabilities (useful for measuring histogram concentration; high values indicate that many pixels share similar intensities)
- hist_entropy_log10: Histogram entropy computed using logarithm base 10.
- (useful as an alternative entropy scale for capturing intensity-distribution complexity)
- log_energy: Sum of log-squared pixel values (useful for compressing the range of histogram-energy values and comparing concentration across datasets)

## B.3 CONTRAST & DYNAMIC RANGE

- robust_contrast_5_95: Intensity range between 5th and 95th percentiles (useful for measuring practical contrast while reducing the influence of extreme outliers)
- michelson_contrast: (max - min) / (max + min) (useful for measuring relative contrast and scanner/post-processing differences in brightness scaling)
- dynamic_range: max - min; full range of intensities (useful for identifying differences in intensity scaling, compression, gain, and clipping)
- shadow_clip_pct: Percentage of pixels clipped at the dark end (useful for detecting black regions, acoustic shadowing, background dominance, or aggressive preprocessing)
- highlight_clip_pct: Percentage of pixels clipped at the bright end (useful for detecting saturation caused by high gain, strong reflectors, or scanner-specific compression)

## B.4 SHARPNESS, NOISE & EDGE

- sharpness_var_laplacian: Variance of the Laplacian-filtered image; higher = sharper (useful for capturing image sharpness, focus, edge strength, and differences in resolution or smoothing)
- noise_estimate: Estimated noise level in the region (useful for detecting dataset differences in acquisition noise, speckle noise, and preprocessing smoothness)
- speckle_contrast: std / mean; classic ultrasound speckle metric (useful for capturing ultrasound speckle behaviour, which can vary across scanner vendors, probes, presets, and image-processing pipelines)
- edge_density: Fraction of pixels classified as edges (useful for capturing structural complexity, border visibility, and differences in sharpness or segmentation difficulty)
- grad_mean: Mean of the image gradient magnitude (useful for measuring average edge strength and local intensity transitions)
- grad_std: Standard deviation of gradient magnitudes (useful for capturing variability in edge strength and structural heterogeneity)
- grad_skew: Skewness of gradient magnitudes (useful for identifying images where a small number of very strong edges dominate)
- grad_kurt: Kurtosis of gradient magnitudes (useful for detecting heavy-tailed edge distributions caused by sharp borders, artefacts, or noise spikes)
- grad_nonzero_pct: Percentage of pixels with non-zero gradient (useful for measuring how much of the region contains visible structural variation rather than flat intensity)

## B.5 GLCM: GREY-LEVEL CO-OCCURRENCE MATRIX

- contrast: Local intensity variation between neighbouring pixels (useful for measuring local texture contrast and scanner-dependent tissue appearance)
- correlation: Linear dependency of grey levels between neighbours (useful for capturing spatial organisation and repeated texture patterns)
- energy / ASM: Angular Second Moment; uniformity of texture (useful for identifying homogeneous regions versus complex or noisy texture)
- homogeneity: Closeness of GLCM to its diagonal; higher = smoother (useful for measuring smoothness; high values may indicate smoother tissue appearance or stronger preprocessing)
- dissimilarity: Weighted contrast between neighbouring pixels (useful for capturing local grey-level differences and texture roughness)
- entropy: Randomness of grey-level pairs (useful for measuring texture complexity, speckle randomness, and image heterogeneity)
- sum_average: Mean of the sum of co-occurring grey levels (useful for capturing average local brightness relationships)
- sum_variance: Variance of the sum distribution (useful for capturing variability in local brightness combinations)
- sum_of_squares: Variance of the GLCM about the mean (useful for measuring spread in grey-level co-occurrence patterns)
- inverse_difference_moment: Smoothness; penalises large grey-level differences (useful for identifying smoother images or images affected by denoising and interpolation)
- glcm_entropy_log10: Log10 of GLCM entropy (useful for capturing local texture complexity on an alternative entropy scale)

## B.6 GLRLM: GREY-LEVEL RUN-LENGTH MATRIX

- SRE: Short Run Emphasis; favours fine texture (useful for detecting fine texture, speckle granularity, and rapid local intensity changes)
- LRE: Long Run Emphasis; favours coarse/uniform regions (useful for detecting coarse texture, smooth areas, or elongated homogeneous structures)

- GLN: Grey Level Non-uniformity across runs (useful for measuring whether run textures are dominated by a few grey levels)
- RLN: Run Length Non-uniformity (useful for measuring whether texture contains consistent or variable run lengths)
- RP: Run Percentage; fraction of possible runs that exist (useful for measuring run density and overall texture granularity)
- LGRE: Low Grey-level Run Emphasis (useful for detecting dark homogeneous runs, which may reflect shadows, background, or low-gain regions)
- HGRE: High Grey-level Run Emphasis (useful for detecting bright homogeneous runs, which may reflect strong reflectors or high-gain settings)
- SRLGE: Short Run Low Grey-level Emphasis (useful for detecting fine dark texture and low-intensity speckle patterns)
- SRHGE: Short Run High Grey-level Emphasis (useful for detecting fine bright texture and high-intensity speckle patterns)
- LRLGE: Long Run Low Grey-level Emphasis (useful for detecting extended dark regions such as shadowing or background)
- LRHGE: Long Run High Grey-level Emphasis (useful for detecting extended bright structures or highly enhanced tissue regions)

## B.7 GLSZM: GREY-LEVEL SIZE ZONE MATRIX

- SZE: Small Zone Emphasis (useful for detecting fine-scale texture and fragmented speckle patterns)
- LZE: Large Zone Emphasis (useful for detecting large homogeneous regions, smoothing, or low-detail areas)
- LGZE: Low Grey-level Zone Emphasis (useful for identifying dark zones caused by background, shadowing, or low gain)
- HGZE: High Grey-level Zone Emphasis (useful for identifying bright zones caused by high gain, strong tissue echoes, or saturation)
- SZLGE: Small Zone Low Grey-level Emphasis (useful for detecting small dark texture components)
- SZHGE: Small Zone High Grey-level Emphasis (useful for detecting small bright speckles or bright local structures)
- LZLGE: Large Zone Low Grey-level Emphasis (useful for detecting large dark homogeneous regions)
- LZHGE: Large Zone High Grey-level Emphasis (useful for detecting large bright homogeneous regions)
- GLNU_zone: Grey Level Non-uniformity for zones (useful for measuring whether zone texture is dominated by particular grey levels)
- ZLNU: Zone Length Non-uniformity (useful for measuring variability in zone sizes and structural regularity)
- ZP: Zone Percentage; fraction of pixels in zones (useful for measuring zone density and texture fragmentation)

## B.8 NGTDM: NEIGHBOURHOOD GREY TONE DIFFERENCE MATRIX

- coarseness: Inverse of average intensity difference with neighbourhood; high = coarse (useful for detecting coarse versus fine image texture)
- contrast_ngtdm: Intensity variation relative to neighbourhood differences (useful for measuring local contrast and tissue-boundary visibility)
- busyness: Rapid spatial intensity changes; high = busy/noisy texture (useful for detecting noisy, speckled, or highly variable local texture)

## B.9 LBP: LOCAL BINARY PATTERN

- lbp_entropy: Entropy of the LBP histogram at radius 1 (useful for capturing local binary texture complexity and speckle pattern variability)
- lbp_uniformity: Fraction of uniform LBP codes at radius 1 (useful for identifying whether local texture patterns are repetitive or diverse)
- lbp_mean: Mean of LBP code values at radius 1 (useful for summarising dominant local binary texture patterns)
- lbp_std: Standard deviation of LBP codes at radius 1(useful for capturing variability in local texture patterns)
- lbp_r2_entropy/uniformity/mean/std: Same metrics at radius 2 (useful for capturing slightly larger-scale texture and speckle structure)
- lbp_r3_entropy/uniformity/mean/std: Same metrics at radius 3 (useful for capturing broader local texture patterns and dataset-specific appearance differences)
- cslbp_r1_entropy/uniformity/mean/std: Centre-Symmetric LBP at radius 1 (useful for capturing local symmetric texture patterns while being more compact than standard LBP)

## B.10 GABOR FILTER FEATURES

- gabor_0_mean/std: Response to horizontal (0°) Gabor filter (useful for capturing directional texture and edge content in one orientation)
- gabor_45_mean/std: Response at 45° orientation (useful for capturing diagonal texture or directional image structures)
- gabor_90_mean/std: Response at vertical (90°) orientation (useful for capturing vertically oriented texture or anatomical boundaries)
- gabor_135_mean/std: Response at 135° orientation (useful for capturing the opposite diagonal orientation and directional tissue texture)

## B.11 WAVELET FEATURES

- wavelet_LL_energy: Energy in the coarse/approximation subband (useful for capturing low-frequency brightness and large-scale appearance differences)
- wavelet_LH_energy: Energy in the horizontal edge subband (useful for capturing vertical edge and texture information)
- wavelet_HL_energy: Energy in the vertical edge subband (useful for capturing vertical edge and texture information)
- wavelet_HH_energy: Energy in the diagonal detail/noise subband (useful for capturing high-frequency texture, speckle, noise, and diagonal details)

## B.12 HOG: HISTOGRAM OF ORIENTED GRADIENTS

- hog4_bin0-3: Gradient direction distribution in 4 bins (90° each) (useful for capturing coarse orientation structure and global edge direction patterns)
- hog8_bin0-7: Gradient direction distribution in 8 bins (45° each) (useful for capturing more detailed directional edge structure)
- hog16_bin0-15: Gradient direction distribution in 16 bins (22.5° each) (useful for capturing fine orientation differences in anatomical boundaries and ultrasound texture)

## B.13 AR MODEL: AUTOREGRESSIVE MODEL

- ar_theta1-4: AR model coefficients; spatial dependency weights (useful for capturing how strongly each pixel relates to its neighbouring pixels, which reflects local smoothness and texture structure)
- ar_sigma2: Residual variance of the AR model fit (useful for measuring unexplained local variation, which may reflect noise, speckle, or irregular texture)

## B.14 IMAGE QUALITY METRICS

- SNR1: Signal-to-Noise Ratio (mean / std) within the region (useful for comparing image quality, noise level, and contrast consistency across datasets)
- EFC: Entropy Focus Criterion; higher = blurrier image (useful for capturing focus/dispersion-related image quality differences, although it should be interpreted cautiously for echocardiography)

## B.15 LV SHAPE FEATURES

- lv_area_px: Area of the LV mask in pixels (useful for capturing LV size and differences in anatomy, view selection, or segmentation scale)
- lv_area_ratio: LV area as a fraction of total image area (useful for comparing relative LV size after resizing and cropping)
- lv_bbox_area_ratio: LV area relative to its bounding box area (useful for capturing how much of the image is occupied by the LV bounding region)
- lv_bbox_w_ratio: LV bounding box width as a fraction of image width (useful for capturing horizontal LV scale and field-of-view differences)
- lv_bbox_h_ratio: LV bounding box height as a fraction of image height (useful for capturing vertical LV scale and anatomical framing differences)
- lv_centroid_x_norm: Normalised x-position of LV centre (0-1) (useful for detecting differences in LV alignment, cropping, and view positioning)
- lv_centroid_y_norm: Normalised y-position of LV centre (0-1) (useful for detecting differences in LV vertical placement and acquisition framing)
- lv_perimeter_px: Perimeter of the LV mask in pixels (useful for capturing LV boundary length and shape complexity)
- lv_circularity: How circle-like the LV shape is (1 = perfect circle) (useful for capturing LV shape compactness and view-dependent anatomical variation)
- lv_eccentricity: Ellipse eccentricity of the LV shape (0 = circle, 1 = line) (useful for capturing elongation of the LV, which may reflect view differences or cardiac phase variation)

## B.16 CROSS-REGION CONTRAST (LV vs NEAR-LV)

- dsmri_lv_vs_nearlv__SNR2: SNR using LV signal relative to near-LV noise (useful for measuring how clearly the LV signal stands out from surrounding tissue)
- dsmri_lv_vs_nearlv__CNR: Contrast-to-Noise Ratio between LV and surrounding tissue (useful for estimating boundary separability and segmentation difficulty)
- dsmri_lv_vs_nearlv__CJV: Coefficient of Joint Variation; lower = better tissue separation (useful for measuring tissue separation; lower values suggest better LV-to-background distinction)

## B.17 Transfer Metadata Columns

- train_dataset: Dataset used as the source training domain. (useful for identifying which domain the segmentation model was trained on)

- test_dataset: Dataset used as the target evaluation domain. (useful for identifying the target-domain where transfer performance is measured)
- dice: Image-level Dice score for the segmentation result. (useful as the main segmentation-performance outcome)
- same_domain: Binary indicator showing whether the training and test datasets are the same. (useful for separating within-domain performance from cross-domain transfer performance)
- source_target_pair: Combined source-to-target label, written as train_dataset__test_dataset. (useful for grouping results by transfer direction, for example CAMUS-to-EchoNet or Unity-to-CAMUS)

## B.18 Individual Feature Distance Columns

To make the transfer predictors directly interpretable, each target image was compared with the source training distribution. For every source dataset and every retained feature, the source mean and standard deviation were computed. The target feature value was then converted into three source-normalised predictors including an absolute z-distance z__<feature>, a signed z-distance sz__<feature>, and a log-compressed absolute distance logz__<feature>. The absolute z-distance measured how far the target image was from the source distribution regardless of direction, the signed z-distance preserved whether the target was above or below the source mean, and the log-compressed version reduced the influence of very large deviations. Z-distances were clipped to avoid extreme propagation.

Feature-family summaries were then created by averaging the individual z-distance predictors within each feature-family. The famz__<family> variables represented mean absolute source-normalised deviation across a feature-family, while famsz__<family> variables represented mean signed deviation across the same family. These family-level predictors allowed the analysis to test whether broader groups such as brightness/contrast, image quality, texture, edge structure, frequency features, geometry, or LV/near-LV context were associated with transfer degradation.

- z__<feature>: Absolute z-distance between the target image feature value and the source training distribution. (useful for measuring how far the target image is from the source training dataset for that feature, measured in source standard-deviation units; for example, a high z__fan__mean means the target image brightness is very different from the source brightness, which may simulate differences in gain, scanner settings, or post-processing)
- logz__<feature>: Log-transformed version of z__<feature>, calculated using log1p(z). (useful for reducing the effect of very large outlier distances while keeping the same interpretation of domain deviation)
- sz__<feature>: Signed z-distance between the target image feature value and the source training distribution. (useful for showing the direction of the domain shift; for example, whether the target image is brighter, darker, sharper, noisier, or more textured than the source images)

The code creates these columns for each original feature. It directly describes z__feat as absolute deviation from the source training distribution, logz__feat as compressed log1p(z), and sz__feat as signed shift direction.

## B.19 Family-Level Distance Columns

- famz__brightness_contrast: Mean absolute z-distance across brightness and contrast features. (useful for measuring overall brightness/gain/contrast shift between source and target domains)
- famz__image_quality_appearance: Mean absolute z-distance across image-quality and appearance features. (useful for measuring differences in noise, entropy, speckle, clipping, and general image appearance)

- famz__texture_radiomics: Mean absolute z-distance across radiomic texture features such as GLCM, GLRLM, GLSZM, and NGTDM. (useful for measuring scanner- or dataset-dependent tissue texture differences)
- famz__local_texture_lbp: Mean absolute z-distance across LBP and CS-LBP features. (useful for measuring local speckle-pattern and micro-texture differences)
- famz__edge_gradient_hog: Mean absolute z-distance across edge, gradient, and HOG features. (useful for measuring differences in boundary sharpness, edge visibility, and anatomical structure clarity)
- famz__frequency_filter_ar: Mean absolute z-distance across Gabor, wavelet, and autoregressive features. (useful for measuring differences in directional texture, frequency content, smoothing, and spatial dependency)
- famz__geometry_framing: Mean absolute z-distance across geometry and framing-related features. (useful for measuring differences in field of view, LV size, image framing, and anatomical positioning)
- famz__lv_nearlv_boundary_context: Mean absolute z-distance across LV-region, near-LV-region, and LV-vs-near-LV features. (useful for measuring whether the LV and surrounding tissue appearance differ from the source-domain)
- famz__other: Mean absolute z-distance across features not assigned to the predefined families. (useful as a residual summary for features not captured by the main semantic groups)

Important note: the exact family columns appear only if that family exists after feature cleaning. The code builds famz__<family> and famsz__<family> automatically from the feature-family map. z__feat as the absolute deviation from the source training distribution, while famz__family is the mean z-distance across all features belonging to the same family.

## B.20 PCA, Mahalanobis, and Nearest-Neighbour Distance Columns

In addition to individual and family-level z-distance features, multivariate image-to-source distance measures were computed. For each source dataset, PCA was fitted on source-standardised features using up to 20 components, depending on the number of available features and samples. Each target image was projected into the corresponding source PCA space. Four distance descriptors were then calculated: Euclidean distance in PCA space, Mahalanobis distance in PCA space, mean nearest-neighbour distance to source-domain samples, and minimum nearest-neighbour distance to the closest source-domain sample. These variables measured whether a target image was globally close to, or far from, the source distribution in multivariate handcrafted-feature space.

- pca_euc: Euclidean distance between one target image and the source training dataset distribution in PCA space. (useful for measuring overall multivariate domain distance after reducing correlated feature dimensions)
- pca_mah: Mahalanobis distance, a statistical measure that calculates the distance between a data point and a distribution, of a target image from the source distribution in PCA space. (useful for measuring how unusual the target image is relative to the covariance structure of the source-domain)
- knn_mean: Mean distance from the target image to its nearest-neighbours in the source PCA space. (useful for measuring how close the target image is to typical source-domain samples)
- knn_min: Minimum nearest-neighbour distance from the target image to the source PCA space. (useful for checking whether the source training set contains at least one image similar to the target image; larger values mean that even the closest source image is dissimilar)

## B.21 Domain-Probability Columns

Out-of-fold domain probabilities were also computed using a balanced logistic-regression domain classifier with median imputation, feature standardisation, and stratified K-fold cross-validation. The classifier produced domainprob__prob_<dataset> columns, indicating how strongly each image resembled each dataset in handcrafted-feature space. For each source-target transfer row, the probability assigned to the source dataset was extracted as domainprob__source_probability, and its complement was stored as domainprob__one_minus_source_probability. These predictors measured how source-like or non-source-like each target image appeared.

- domainprob__prob_<dataset>: Out-of-fold probability that an image resembles a specific dataset according to a balanced logistic-regression domain classifier with median imputation, feature standardisation, and StratifiedKFold cross-validation. (useful for measuring how strongly an image resembles each dataset/domain based on handcrafted features)
- domainprob__source_probability: Domain-classifier probability assigned to the source dataset. (useful for measuring how source-like the target image appears)
- domainprob__one_minus_source_probability: One minus the source-domain probability. (useful for measuring how non-source-like the target image appears; higher values indicate stronger apparent domain mismatch)

The code creates out-of-fold dataset probabilities and then derives source probability and one-minus-source probability for transfer analysis

## B.22 Source Baseline Performance Columns

- within_source_dice_mean: Mean Dice score when the model is evaluated within the same source-domain. (useful for estimating the baseline strength of the source model before cross-domain transfer)
- within_source_dice_std: Standard deviation of within-source Dice scores. (useful for measuring how stable or variable the source-domain performance is)
- within_source_n: Number of within-source samples used to estimate the source baseline. (useful for checking reliability of the within-source baseline estimate)
- srcbase__within_source_dice_mean: Copied source-baseline mean Dice used as a model predictor. (useful because stronger source-domain performance may influence cross-domain transfer performance)
- srcbase__within_source_dice_std: Copied source-baseline Dice standard deviation used as a model predictor. (useful because unstable source-domain performance may indicate poorer generalisation)

## B.23 Source Dataset Size Columns

- srcsize__train_n: Number of training images in the source dataset. (useful because larger source training sets may improve robustness and generalisation)
- srcsize__log_train_n: Log10-transformed number of source training images. (useful for representing source dataset size on a compressed scale, because dataset sizes can differ greatly)
- srcsize__val_n: Number of validation images in the source dataset. (useful for describing the amount of validation data available for source-domain model development)
- srcsize__test_n: Number of test images in the source dataset. (useful for describing the source dataset scale and evaluation reliability)
- srcsize__total_n: Total number of images in the source dataset. (useful for measuring overall dataset size as a source-quality indicator)

## B.24 Extended Source/Domain Distance Columns

- srcdist__zscore_rms: Root-mean-square z-distance across all features relative to the source distribution. (useful as a single global measure of how far the target image is from the source-domain across all handcrafted features)
- srcdist__mahalanobis_lw: Ledoit-Wolf Mahalanobis distance in PCA space relative to the source distribution. (useful for measuring multivariate domain abnormality with covariance regularisation)

- srcdist__nearest_other_mahalanobis: Mahalanobis distance to the nearest non-source-dataset distribution. (useful for identifying whether the target image is closer to another dataset than to the source)
- srcdist__second_nearest_other_mahalanobis: Mahalanobis distance to the second-nearest non-source-dataset distribution. (useful for measuring how many alternative domains the target resembles)
- srcdist__dataset_ambiguity: Ratio between the second-nearest and nearest non-source-dataset distances. (useful for measuring whether the target image clearly resembles one alternative dataset or is ambiguous between multiple datasets)
- srcdist__mean_other_mahalanobis: Mean Mahalanobis distance to all non-source-dataset distributions. (useful for measuring overall separation from the rest of the dataset population)

### B.25 Target Outcome Columns

- target__dice: Dice score used as the direct prediction target. (useful for modelling absolute segmentation performance)
- target__drop_from_source_mean: Difference between the source within-domain mean Dice and the cross-domain Dice score. (useful for modelling domain-transfer degradation; higher values mean larger performance drop)
- target__relative_drop_from_source_mean: Relative Dice drop, calculated as drop from source mean divided by within-source mean Dice. (useful for normalising degradation by the strength of the original source model)

## Appendix C. Modelling Details

### C.1 Variational Autoencoder

The VAE used in this study adopted a convolutional encoder-decoder architecture designed for greyscale echocardiographic images resized to 224×224 pixels. The latent space dimension was set to 256.

The encoder consisted of five convolutional blocks with 4×4 kernels, stride 2, and padding 1, progressively increasing the number of feature channels from 1 to 32, 64, 128, 256, and 512, whilst reducing the spatial resolution from 224×224 to 7×7. Each block was followed by batch normalisation and a LeakyReLU activation ($\alpha = 0.2$). The final feature map was flattened and mapped to two fully connected layers that estimated the mean ($\mu$) and log-variance of the latent distribution. Latent samples were obtained using the reparameterisation trick.

The decoder mirrored the encoder through a fully connected projection followed by five transposed convolution blocks that progressively reconstructed the image to the original resolution. The final layer consisted of a 3×3 convolution followed by a sigmoid activation function to produce the normalised output image.

Training was performed using the Adam optimiser with a learning rate of $10^{-4}$, weight decay of $10^{-6}$, and a batch size of 8 for a maximum of 80 epochs. A KL warm-up strategy was applied, with the regularisation weight ($\beta$) increasing linearly from 0 to 0.003 over the first 30 epochs. Model selection and early stopping were based on validation reconstruction loss.

The reconstruction objective consisted of an L1 reconstruction loss combined with KL divergence regularisation. No additional edge-preserving loss terms were included in the final configuration, as they did not improve validation performance.

The same preprocessing pipeline described for the segmentation model, including mask-aware cropping and resizing, was applied to ensure consistency across representation learning frameworks. A three-seed sensitivity analysis of RVENet inclusion in VAE training is reported in Appendix H.

## C.2 LV Segmentation Model

The LV segmentation model was based on a U-Net architecture with an encoder–decoder structure and skip connections. The encoder consisted of five hierarchical stages, each comprising a convolutional block followed by max-pooling for spatial downsampling. Each convolutional block included two consecutive 3×3 convolution layers, each followed by batch normalisation and ReLU activation. The number of feature channels increased progressively from 32 to 512 across the encoder. At the bottleneck, a central convolutional block expanded the representation to 1024 channels, forming a high-capacity latent representation of the input image.

The decoder mirrored the encoder through a sequence of upsampling stages implemented using transposed convolutions. At each stage, the upsampled feature maps were concatenated with the corresponding encoder feature maps via skip connections, followed by convolutional processing with batch normalisation and ReLU activation. A final 1×1 convolution layer mapped the decoder features to a single-channel output, and a sigmoid activation function was applied to produce a pixel-wise probability map for binary segmentation.

The network operated on single-channel greyscale images and was trained using the Adam optimiser with an initial learning rate of $10^{-4}$ and a batch size of 16. Training was performed for a maximum of 200 epochs. Early stopping was applied with a patience of 12 epochs and a minimum improvement threshold of 0.001. A ReduceLROnPlateau scheduler was used to adapt the learning rate based on validation performance, reducing the learning rate by a factor of 0.5 when no improvement was observed for 5 consecutive epochs, with a minimum learning rate of $10^{-6}$. The loss function was binary cross-entropy, and model performance was monitored using Intersection over Union (IoU) and Dice similarity coefficient on the validation set.

Preprocessing included fan-shaped ROI masking to remove non-informative regions, followed by mask-aware cropping and a secondary tight cropping step to isolate the valid imaging area. The resulting images were resized to a fixed resolution of 224×224 pixels using aspect-ratio-preserving resizing with zero-padding. Intensity normalisation was applied to standardise input distributions across datasets. No additional data augmentation was used during training.

In addition to segmentation outputs, latent features were extracted from the bottleneck layer of the U-Net. These features captured high-level anatomical and structural information learned during training. To enable their use in downstream analysis, the bottleneck feature maps were transformed into fixed-length feature vectors using global average pooling. These vectors were used as task-specific latent representations for domain shift analysis and comparison with image-based latent features derived from the VAE.

## Appendix D. Image Classification Ablation

To further investigate the role of superficial image cues in domain separation, we conducted an image-classification ablation study. Models such as VGG16, as well as smaller CNN architectures, achieved high accuracy in distinguishing between datasets by exploiting superficial image cues, including cropping, framing, and acquisition-related artefacts, rather than learning domain-invariant representations. Grad-CAM analysis supported this observation, showing that model attention was often directed towards irrelevant regions instead of clinically meaningful structures. This behaviour indicated shortcut learning, where models relied on trivial features that did not generalise across domains. Consequently, high classification performance did not imply true domain understanding or similarity, as shown in Fig. 17.

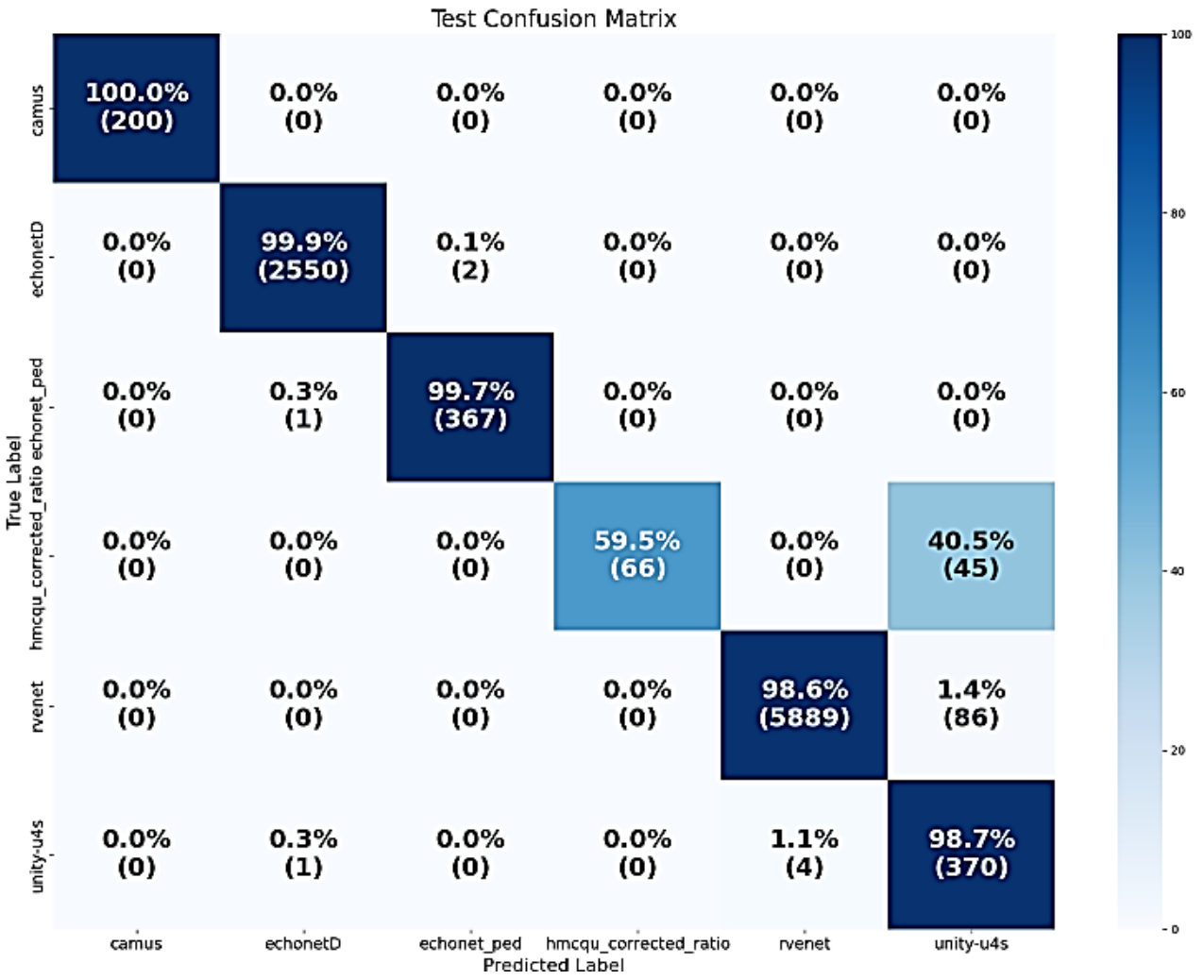

*Fig. 17. Confusion matrix for dataset-classification ablation*

## Appendix E. Segmentation Preprocessing Ablation

Table 11 reports the Dice scores obtained across all datasets when LV masks were used without any preprocessing. Table 12 presents the results obtained after applying padding followed by resizing, without cropping. Compared with these settings, the proposed preprocessing approach generally improved cross-dataset performance, particularly in several poor transfer cases. This improvement was attributed to its ability to preserve the fan-shaped ultrasound geometry, maintain the structural integrity of the LV region, and produce a more standardised image representation in terms of near-field and far-field appearance relative to the probe/viewpoint. By preserving this spatial relationship, the model could better learn the anatomical position and shape of the LV cavity across datasets.

*Table 11. Cross-dataset Dice scores without preprocessing*

| Train | UN | CS | CA | ED | EP | HMC-QU |
|---|---|---|---|---|---|---|
| UN | 0.9212 | 0.9342 | 0.5990 | 0.7192 | 0.6646 | 0.8396 |
| CA | 0.4870 | 0.4981 | 0.9365 | 0.5446 | 0.6867 | 0.6378 |
| ED | 0.7192 | 0.7928 | 0.7436 | 0.9318 | 0.8673 | 0.8688 |
| EP | 0.5669 | 0.5836 | 0.7952 | 0.8728 | 0.8954 | 0.8006 |
| HMC-QU | 0.7376 | 0.6424 | 0.2425 | 0.2957 | 0.3248 | 0.8906 |

***Table 12. Cross-dataset Dice scores after padding and resizing only***

| Train | UN | CS | CA | ED | EP | HMC-QU |
|---|---|---|---|---|---|---|
| UN | 0.9212 | 0.9342 | 0.7044 | 0.7192 | 0.6646 | 0.9066 |
| CA | 0.5218 | 0.5384 | 0.9347 | 0.4854 | 0.6407 | 0.6745 |
| ED | 0.7192 | 0.7928 | 0.7160 | 0.9318 | 0.8673 | 0.8370 |
| EP | 0.5669 | 0.5836 | 0.7833 | 0.8728 | 0.8954 | 0.7932 |
| HMC-QU | 0.7376 | 0.6424 | 0.2835 | 0.2957 | 0.3248 | 0.9466 |
| All | 0.9200 | 0.9338 | 0.9356 | 0.9311 | 0.8973 | 0.9466 |

Table 13 suggests that cross-vendor generalisation may appear stronger than several of the cross-dataset transfers reported in Table 3. However, this comparison should be interpreted with caution. The apparent improvement may be influenced by partial overlap in vendor characteristics across datasets, particularly the presence of GE images in multiple sources. Therefore, the model may learn shared acquisition- or dataset-specific features rather than scanner-vendor characteristics alone. For this reason, the main evaluation in this study is based primarily on the cross-dataset results reported in Table 3.

***Table 13. Cross-vendor LV segmentation Dice scores***

| Train | GE | PH |
|---|---|---|
| GE | 0.9299 | 0.9071 |
| PH | 0.8228 | 0.9278 |
| All | 0.9298 | 0.9274 |

# Appendix F. Feature-importance Analysis

Dataset-level feature importance was computed for the target drop from source mean using an HGB model (as the best model) with the full normalised predictor set under LOSTPO validation. The corresponding model performance was $R^2 = 0.612$. Table 14 ranks the top 20 important features.

***Table 14. Feature importance for the normalised all-feature transfer model, 143 features setting, 25 folds***

| Feature | Perm Mean |
|---|---|
| z__lv__LRE | 0.2598 |
| z__lv__LGRE | 0.1119 |
| z__lv__busyness | 0.0984 |
| z__lv_centroid_x_norm | 0.0972 |
| z__lv__RP | 0.0457 |
| famsz__geometry_framing | 0.0315 |
| famsz__lv_nearlv_boundary_context | 0.0282 |
| z__lv__lbp_uniformity | 0.0278 |
| z__fan__log_energy | 0.0251 |
| famz__lv_nearlv_boundary_context | 0.0236 |
| z__whole__lbp_r3_std | 0.0235 |
| domainprob__prob_consensus | 0.0213 |
| famz__geometry_framing | 0.0205 |
| z__lv__RLN | 0.0182 |
| famz__frequency_filter_ar | 0.0151 |
| domainprob__prob_unity-u4s | 0.0142 |
| z__whole__median | 0.0137 |
| z__fan__cslbp_r1_entropy | 0.0120 |
| srcdist__second_nearest_other_mahalanobis | 0.0117 |
| z__lv__kurtosis_excess | 0.0102 |

# Appendix G. Predictor-block Ablation

Table 15 defines the predictor groups used to interpret the analysis.

***Table 15. Predictor-block definitions used in the ablation analysis***

| Predictor group | Representative variables | Interpretation |
|---|---|---|
| A. Source baseline performance and dataset scale | srcbase__within_source_dice_mean, srcbase__within_source_dice_std, srcsize__train_n, srcsize__log_train_n, srcsize__total_n | Describes source-model strength, source-domain stability and source dataset scale. |
| B. Source-to-target distance | pca_euc, pca_mah, knn_mean, knn_min, srcdist__zscore_rms, srcdist__mahalanobis_lw, srcdist__nearest_other_mahalanobis | Measures how far each target image lies from the corresponding source training distribution in multivariate feature space. |
| C. Individual feature deviations | Training-fold selected z__<feature> and logz__<feature> variables | Captures specific source-normalised deviations in individual image features. The selected features and K value are chosen inside the training folds. |
| D. Family-level summaries | famz__<family> and famsz__<family> variables | Summarises source-normalised deviations across interpretable families such as brightness/contrast, image quality, texture/radiomics, local LBP, edge/HOG, frequency and geometry/framing. |
| E. Domain-probability descriptors | domainprob__prob_<dataset>, domainprob__source_probability, domainprob__one_minus_source_probability | Measures how source-like or non-source-like each target image appears according to the handcrafted-feature domain classifier. |

For Dataset-level, all 31 non-empty subsets of the five blocks were evaluated under the same model and validation scheme. The top ten subsets by $R^2$ are reported in Table 16. The full five-block model attained $R^2 = 0.61$, but the four-block model that drops the domain-probability block reached $R^2 = 0.63$, the highest of any subset. The two-block combination B + C alone already reached 0.59, only 0.015 below the full model, indicating that image-to-source distances together with mRMR-selected per-feature deviations carry the bulk of the predictable signal. As shown in Fig. 18, the block-attribution analysis indicates that the selected C-block z-distance and log-z-distance features and B-block distance descriptors made the strongest contributions to the normalised-all HGB model, while the E-domain-probability block contributed little or negatively.

***Table 16. Top ten predictor-block subsets ranked by cross-validated $R^2$ (HGB, LOSTPO, K = 50).***

| Block set | Blocks | n pred. | $R^2$ | MAE | Spearman ρ |
|---|---|---|---|---|---|
| A + B + C + D | 4 | 135 | 0.632 | 0.0787 | 0.730 |
| B + C + D | 3 | 128 | 0.626 | 0.0792 | 0.732 |
| A + B + C + D + E | 5 (full) | 143 | 0.611 | 0.0821 | 0.681 |
| A + B + C | 3 | 117 | 0.607 | 0.0807 | 0.725 |
| A + C + D | 3 | 125 | 0.597 | 0.0828 | 0.700 |
| B + C | 2 | 110 | 0.596 | 0.0814 | 0.736 |
| B + C + D + E | 4 | 136 | 0.585 | 0.0861 | 0.629 |
| A + B + D | 3 | 35 | 0.582 | 0.0855 | 0.682 |
| C + D | 2 | 118 | 0.580 | 0.0851 | 0.696 |
| A + C + D + E | 4 | 133 | 0.578 | 0.0867 | 0.620 |

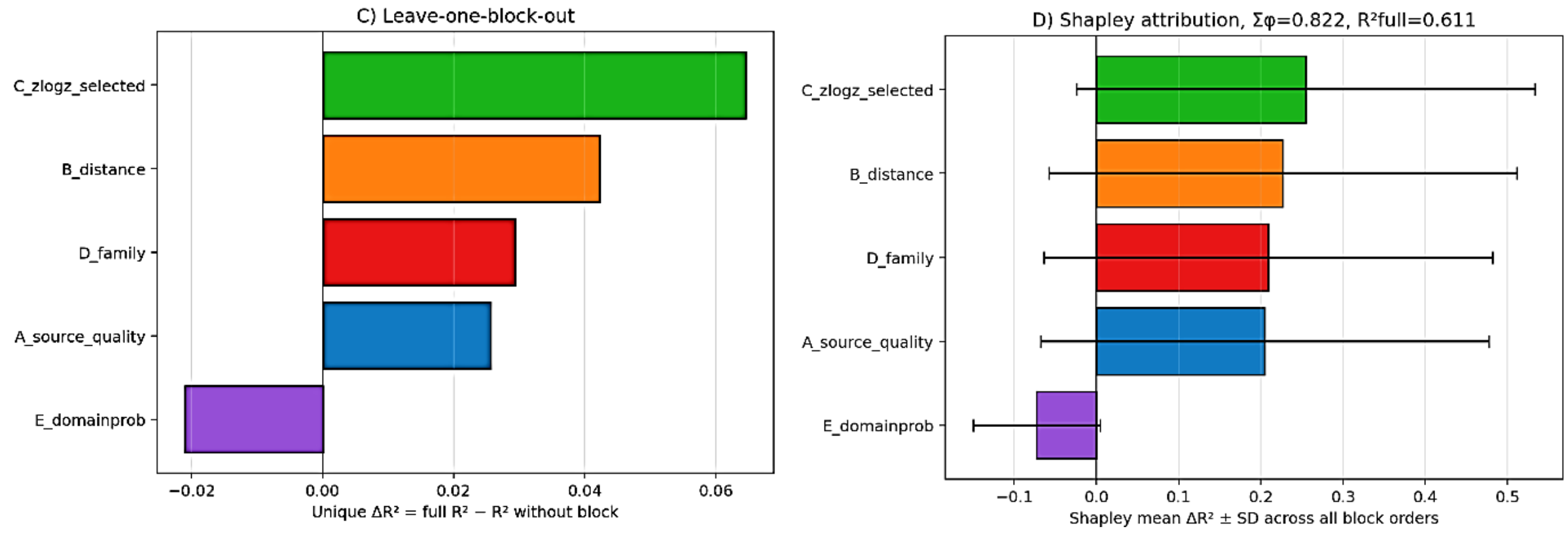

A B

***Fig. 18. Dataset-level block-attribution diagnostic for the normalised-all configuration (A) Leave-one-block-out unique $\Delta R^2$ (B) Top-level Shapley attribution with across orders standard deviation***

# Appendix H. Supplementary Evaluation of Domain-Discrepancy Measures

Fig. 19 provides qualitative VAE discrepancy maps. The $MMD^2$ representation shows clearer separation of CAMUS from the other datasets, whereas the CMD representation corresponds less closely to the observed transfer-performance pattern.

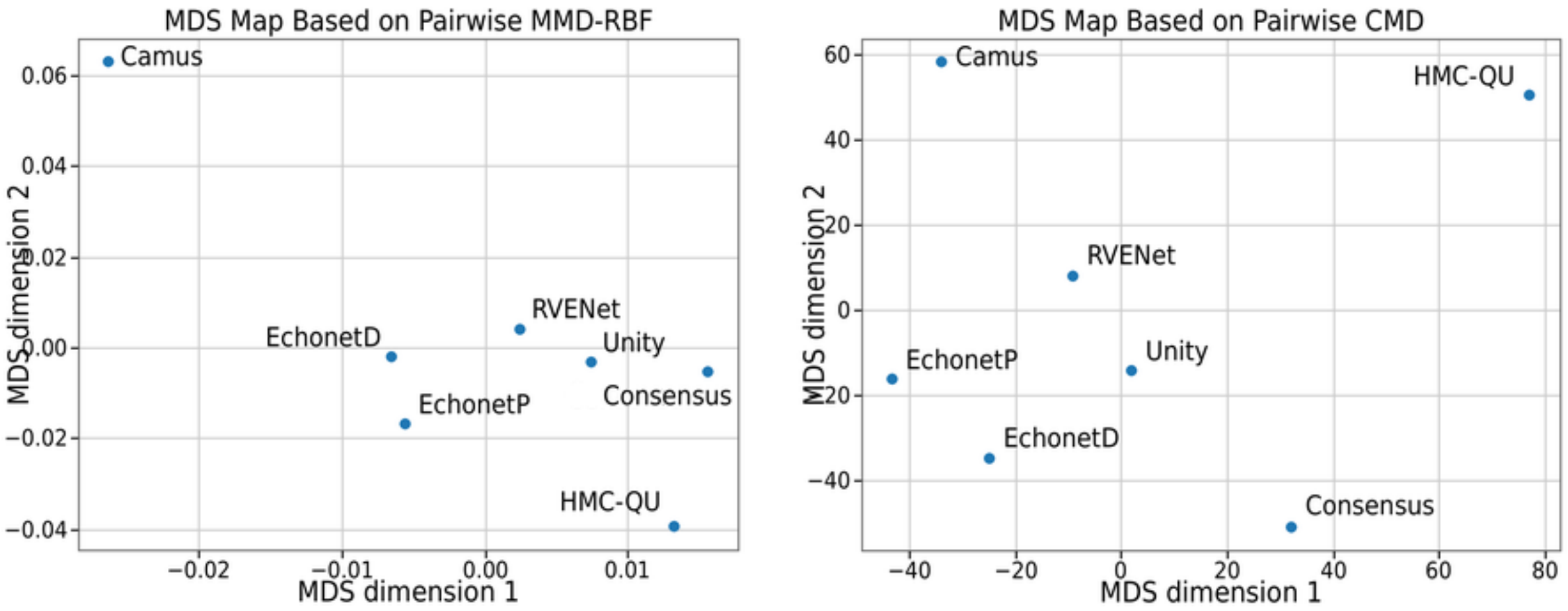

***Fig. 19. VAE latent-space discrepancy maps. (A) MMD² (B) CMD***

Moreover, Fig. 20 shows that MMD² is negatively associated with pairwise Dice in both the VAE and LV-segmentation latent spaces. This relationship is stronger and more consistent in the LV-segmentation space, in agreement with its higher correlation and explained variance. For the retained JS and KL measures, a sensitivity analysis was conducted using 32, 64, and 128 histogram bins to assess the influence of bin selection on the discrepancy-performance correlations. The correlations with pairwise Dice changed only slightly for VAE and LV-segmentation latent features (≤0.002) and the main handcrafted configuration (≤0.04). However, some other handcrafted configurations showed greater sensitivity to bin selection.

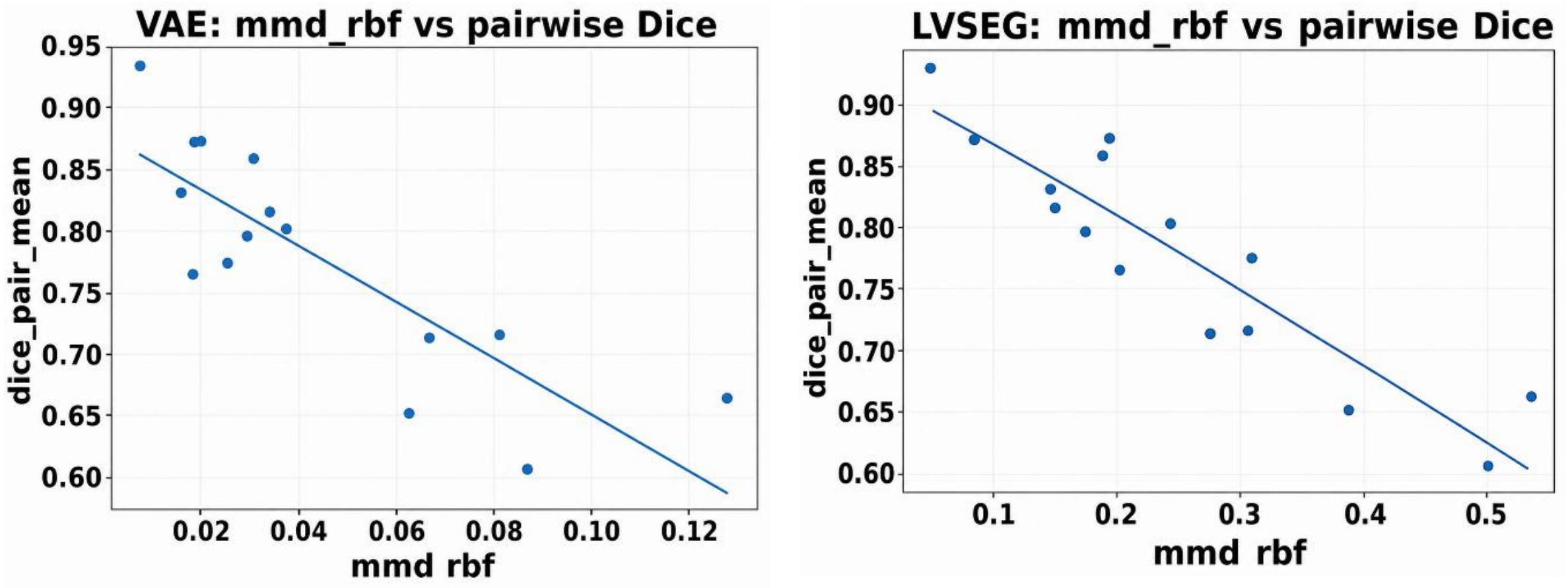

***Fig. 20. Relationship between MMD² and Dice score in VAE and LV-segmentation latent features***

Table 17 presents the pair-LOO regression results across the handcrafted and latent representations. The relationship between dataset discrepancy and Dice performance depended on both the feature representation and discrepancy measure. MMD² performed best among the raw handcrafted configurations, while CMD performed best for normalised handcrafted features. W provided the strongest results for both latent representations. The log1p transformation had relatively little effect on W and MMD² in most configurations, whereas CMD was more sensitive to this transformation.

***Table 17. Pair-LOO $R^2$ for different feature representations and discrepancy measures.***

| Representation | W | MMD² | CMD |
|---|---|---|---|
| Raw All | 0.184 / 0.217 | 0.337 / 0.338 | -0.297 / -0.881 |
| Raw No LV | 0.175 / 0.215 | 0.360 / 0.355 | -0.263 / -0.199 |
| Raw Whole | 0.065 / 0.096 | 0.246 / 0.247 | -0.258 / -0.210 |
| Norm. All | 0.215 / 0.243 | 0.295 / 0.300 | 0.651 / 0.009 |
| Norm. No LV | 0.330 / 0.367 | 0.407 / 0.402 | 0.648 / 0.588 |
| Norm. Whole | 0.430 / 0.438 | 0.424 / 0.407 | 0.640 / 0.620 |
| VAE | 0.590 / 0.590 | 0.463 / 0.489 | -0.126 / -0.088 |
| LV-seg | 0.852 / 0.846 | 0.748 / 0.777 | -0.321 / -0.320 |

**Note.** Each cell reports raw / log1p-transformed discrepancy results. All regressions used the original 15 dataset pairs. Raw and Norm. refer to unnormalised and normalised handcrafted features; No LV excludes LV-related descriptors, and Whole contains whole-image features only.

To assess the influence of individual datasets, each of the six datasets was excluded in turn, leaving ten pairs. We also examined all 3,003 possible ten-pair subsets to determine whether similar changes could arise from removing other combinations of five pairs. As shown in Table 18, excluding CAMUS substantially weakened both the correlations and pair-LOO performance for several representations, particularly the latent and normalised handcrafted features. These changes were unusual compared with most other ten-pair subsets, although the effect was not consistent across all configurations. In some cases, correlation strength and predictive performance also changed differently, highlighting the importance of considering both measures.

***Table 18. Selected dataset-exclusion sensitivity results using raw discrepancy measures.***

| Representation | Pearson r | Pair-LOO R² | Other exclusions R² range | CAMUS percentile (%) |
|---|---|---|---|---|
| Raw All, W | -0.588 to -0.693 | 0.184 to 0.139 | -0.071 to 0.226 | 63.45 |
| Raw All, MMD² | -0.694 to -0.571 | 0.337 to -0.129 | 0.109 to 0.431 | 2.85 |
| Norm. Whole, W | -0.754 to -0.579 | 0.430 to -0.144 | 0.219 to 0.472 | 0.48 |
| Norm. Whole, CMD | -0.852 to -0.001 | 0.640 to -0.637 | 0.455 to 0.746 | 0.75 |
| VAE, W | -0.830 to -0.531 | 0.590 to -0.156 | 0.414 to 0.669 | 0.05 |
| LV-seg, W | -0.940 to -0.765 | 0.852 to 0.371 | 0.817 to 0.891 | 0.12 |

**Note.** For Pearson r and pair-LOO $R^2$, the first value represents the full 15-pair analysis and the second represents the analysis without CAMUS. The other-exclusion ranges cover the five remaining dataset exclusions. CAMUS percentile ranks are based on pair-LOO $R^2$ across all 3,003 ten-pair subsets. Lower percentiles indicate weaker predictive performance. These rankings are descriptive and do not indicate statistical significance.

CAMUS contributed the five lowest-Dice pairs, and its exclusion reduced the standard deviation of Dice outcomes from 0.093 to 0.052. This reduction in outcome variability, together with changes in the range of discrepancy values and the influence of individual pairs, may partly explain the weaker associations. The effect should therefore not be interpreted as evidence that CAMUS necessarily has a uniquely larger domain shift.

The reduction in pair-LOO $R^2$ also did not always indicate worse absolute prediction error relative to a simple baseline. For example, after excluding CAMUS, VAE raw W achieved a pair-LOO MAE of 0.0448, compared with 0.0466 for the cross-validated mean-Dice baseline. The findings remain exploratory because dataset pairs share datasets, the preprocessing was fitted globally, and the overlapping subsets do not provide independent validation or significance tests.

Moreover, our unsupervised VAE included RVENet in its training and validation pools, although RVENet lacked LV segmentation masks and was excluded from the 15-pair Dice analysis. To assess this choice, we first recomputed the original VAE discrepancies using a scaler fitted only on the six LV datasets. We then repeated training with and without RVENet using three seeds (42, 43 and 44), the same architecture and preprocessing, matched optimisation steps per epoch, and model selection based only on validation reconstruction loss. Latent features from the retrained models were standardised using the same images from the six LV datasets, and associations were evaluated against the unchanged pairwise Dice outcomes. As shown in Table 19, changing the feature scaler had negligible effect, while retraining with RVENet reproduced the original VAE results across three seeds.

***Table 19. Sensitivity of VAE discrepancy-Dice associations to RVENet inclusion***

| Measure | Original VAE (A) | Scaler only (B) | With RVENet (A′) | Without RVENet (C) |
|---|---|---|---|---|
| W, Pearson r | -0.830 | -0.830 | -0.831 ± 0.008 | -0.760 ± 0.020 |
| W, pair-LOO $R^2$ | 0.590 | 0.590 | 0.592 ± 0.018 | 0.457 ± 0.044 |
| $MMD^2$, Pearson r | -0.830 | -0.829 | -0.831 ± 0.004 | -0.813 ± 0.007 |
| CMD, Pearson r | -0.332 | -0.365 | -0.391 ± 0.084 | -0.346 ± 0.093 |
| JS, Pearson r | -0.835 | -0.837 | -0.840 ± 0.005 | -0.843 ± 0.002 |
| KL, Pearson r | -0.834 | -0.838 | -0.839 ± 0.005 | -0.843 ± 0.002 |

**Note.** A represents the original VAE trained with RVENet and standardised across seven datasets. B uses the original model with standardisation restricted to the six LV datasets. A′ and C represent retraining with and without RVENet, respectively, using matched seeds 42, 43 and 44. Values for A′ and C are mean ± standard deviation across three seeds. All discrepancy measures are raw, JS and KL use 64 bins, and associations are evaluated on the same 15 dataset pairs.

Removing RVENet weakened the linear association between W and pairwise Dice, whereas $MMD^2$ changed less and JS and KL remained largely unchanged. The W-based Spearman correlation was also similar with and without RVENet. Reconstruction loss on the common six-dataset test set increased by approximately 1% without RVENet. However, because removing RVENet also reduced the number of distinct training images, its specific contribution cannot be isolated. These comparisons remain descriptive given the three seeds and non-independent dataset pairs.

# Appendix I. Secondary Datasets without LV Masks

The RVENet dataset did not provide LV segmentation masks and was therefore excluded from analyses that require LV annotations for the main pipeline [35], [36]. For consistency in temporal sampling, frames from RVENet cine loops were uniformly subsampled at fixed intervals (every 15 frames). Although RVENet does not provide LV segmentation masks, it was retained in the image-level domain-discrepancy analysis because scanner/vendor-related appearance was one of the central questions of this study. In addition, RVENet images were acquired in the apical four-chamber (A4C) view, which is mostly consistent with the primary imaging view used in the other datasets. This ensured that the anatomical perspective and acquisition geometry remained comparable, allowing meaningful analysis of appearance and distributional differences.

MIMIC IV ECHO Ext LVVOLUMES A4C ROI was used as an additional secondary dataset because it provided 1,064 A4C echocardiography cine sequences with manually annotated ROI masks, but not full LV segmentation masks. Since no official train/validation/test split was provided, it was treated only as an external test-style dataset in this study [37]. The characteristics and dataset distributions of RVENet and MIMIC IV ECHO are summarised in Table 20.

***Table 20. Secondary datasets without full LV segmentation masks***

| Features | RVENet | MIMIC IV ECHO Ext LVVOLUMES A4C ROI |
|---|---|---|
| Access | Public | Public |
| Train | 22796 | 0 |
| Validation | 6411 | 0 |
| Test | 5975 | 1064 |
| Total size | 35,182 | 1064 |
| Scanner | Ph - GE | GE |
| Views | A4C | A4C |
| ROI Masks | 60 | 1,064 |

## Appendix J. CycleGAN Experiment

To mitigate domain shift between source and target echocardiographic datasets, we employed a two-stage framework that combined unpaired image translation using CycleGAN with consistency-constrained segmentation learning. A CycleGAN model was first trained on unpaired source and target images to learn bidirectional mappings between domains. The generators followed a ResNet-based architecture with residual blocks and instance normalisation, whilst PatchGAN discriminators were used to enforce local realism. Training was performed using a least-squares adversarial loss, combined with cycle-consistency and identity losses to preserve structural content and stabilise the translation process.

After training, the learned generator was used to translate source-domain images into target-style images. Segmentation masks were not translated; instead, they were resized using nearest-neighbour interpolation to remain spatially aligned with the generated images. This resulted in a synthetic dataset composed of target-style images paired with anatomically accurate source masks, preserving label fidelity whilst adapting image appearance.

In the second stage, a U-Net-based segmentation model was trained using both real source images and their corresponding translated (target-style) counterparts. For each sample, the model processed the real and translated images under supervision of the same ground-truth mask. The training objective combined segmentation supervision on both real and translated images together with a consistency constraint that enforced agreement between predictions obtained from the two domains. The segmentation loss was defined as a weighted combination of binary cross-entropy and Dice loss, whilst the consistency term was implemented as the mean squared error between the predicted probability maps. The overall objective was formulated as a weighted sum of these components, with equal weights assigned to the segmentation losses and a lower weight assigned to the consistency term (set to 0.2).

The segmentation model was trained exclusively using source-domain annotations, including both original and translated images, and evaluated on real target-domain validation and test sets without any fine-tuning. This design allowed assessment of whether enforcing prediction consistency across appearance-shifted representations improved generalisation under domain shift.

Moreover, CycleGAN did not prove to be particularly useful in our experiments. Although its outputs visually resembled the target-domain, the model introduced artificial structures that are not physically plausible in echocardiographic imaging, raising concerns about clinical validity. In addition, it did not contribute to improved segmentation performance or effectively reduce the domain gap in our setting.

# Appendix K. VAE-Reconstructed Images for Segmentation

A separate experiment using a Variational Autoencoder for feature extraction showed that reconstructed images were smoother and exhibited reduced noise; however, segmentation on these reconstructed images did not yield any improvement over using the original inputs. In contrast, the fan-shape mask-aware crop-and-resize strategy with padding demonstrated a consistent positive trend, likely because it preserves anatomical geometry and avoids distortion during preprocessing.